\documentclass[preprint,12pt]{elsarticle}

\usepackage[utf8]{inputenc}
\usepackage{geometry}
\usepackage{amssymb}
\usepackage{amsmath,amssymb,amsfonts}
\usepackage{graphicx}
\usepackage{siunitx}
\usepackage{textcomp}
\usepackage{float}
\usepackage{algorithm}
\usepackage{algpseudocode}
\usepackage{xcolor}
\usepackage{tabularx}
\usepackage{pdfpages}
\usepackage{caption} 
\usepackage{subcaption}
\usepackage{array}
\usepackage{multirow}
\usepackage{orcidlink}
\usepackage{hyperref}
\usepackage{mathrsfs}
\usepackage{booktabs}
\usepackage{bbm}
\usepackage{lipsum}
\usepackage{rotating} 
\usepackage{pdflscape} 

\journal{Elsevier}

\begin{document}

\begin{frontmatter}

\title{Adaptive Real-Time Multi-Loss Function Optimization Using Dynamic Memory Fusion Framework: A Case Study on Breast Cancer Segmentation}

\author{Amin Golnari\corref{cor}\fnref{inst1} \text{\orcidlink{0000-0002-3841-0588}}}
\cortext[cor]{\textit{Corresponding author}}
\ead{amingolnarii@gmail.com}

\author{Mostafa Diba\fnref{inst1} \text{\orcidlink{0000-0003-4198-6112}}}
\ead{m.diba@shahroodut.ac.ir}

\affiliation[inst1]{organization={Faculty of Electrical Engineering, Shahrood University of Technology},
      city={Shahrood},
      country={Iran}}

\begin{abstract}
Deep learning has proven to be a highly effective tool for a wide range of applications, significantly when leveraging the power of multi-loss functions to optimize performance on multiple criteria simultaneously. However, optimal selection and weighting loss functions in deep learning tasks can significantly influence model performance, yet manual tuning of these functions is often inefficient and inflexible. We propose a novel framework called dynamic memory fusion for adaptive multi-loss function penalizing in real-time to address this. This framework leverages historical loss values data to dynamically adjust the weighting of multiple loss functions throughout the training process. Additionally, this framework integrates an auxiliary loss function to enhance model performance in the early stages. To further research horizons, we introduce the class-balanced dice loss function, designed to address class imbalance by prioritizing underrepresented classes. Experiments on breast ultrasound datasets demonstrate that the framework improves segmentation performance across various metrics. These results demonstrate the effectiveness of our proposed framework in ensuring that the model dynamically adjusts its focus to prioritize the most relevant criteria, leading to improved performance in evolving environments. The source code for our proposed methodology is publicly available on GitHub.

\end{abstract}

\end{frontmatter}

\section{Introduction}

Deep learning has revolutionized numerous domains, with medical imaging being one of the most transformative areas. Tasks such as segmentation, classification, and prediction have seen significant advancements due to sophisticated deep learning techniques. However, a key challenge remains, optimizing loss functions to guide model training effectively. Loss functions are critical in determining model performance, yet traditional approaches often rely on manually assigning fixed weights to multiple loss components; a process that is both labor-intensive and suboptimal in dynamic training environments. This issue is further compounded by class imbalance in medical imaging datasets, where underrepresented classes are frequently overlooked, leading to biased learning outcomes. Addressing these challenges requires an innovative approach that dynamically adjusts loss function weights to enhance model adaptability and performance.

This study addresses the pressing need for improved multi-loss function optimization in real-time scenarios, particularly in complex tasks like breast cancer segmentation. Current frameworks often struggle with adaptability when faced with evolving datasets. To overcome these limitations, we propose the Dynamic Memory Fusion (DMF) framework, which adaptively adjusts the weighting of multiple loss functions based on historical data. Additionally, we introduce a class-balanced dice loss function specifically designed to mitigate class imbalance, ensuring that underrepresented classes receive adequate attention during training. By tackling these challenges, our approach aims to provide a robust solution for real-time medical image segmentation, ultimately enhancing diagnostic accuracy and efficiency.

Dynamic weighting of loss functions are crucial for optimizing performance across diverse tasks. This is especially important when addressing challenges such as class imbalance, boundary precision, and feature learning. For instance, GBE-Net \citep{feng2024gbe} demonstrates the benefits of combining cross-entropy and dice loss to improve segmentation accuracy, particularly in preserving lesion boundaries. Similarly, MFMSNet \citep{wu2024mfmsnet} integrates dice and binary cross-entropy losses to address class imbalance while refining tumor edges through multi-scale fusion.

Despite these advances, many studies still rely on fixed-weight schemes for simplicity and consistency, such as those proposed in \citep{zhu2024brain, zhu2024lightweight, zhu2024sparse, zhu2023brain, zhu2025dual}. While these methods achieve reasonable performance by maintaining static weights throughout training, they often fail to adapt to varying complexities or imbalances within datasets. This limitation underscores the need for more flexible approaches.

Dynamic weighting strategies have emerged as a promising solution. For example, SoftAdapt \citep{heydari2019softadapt} dynamically adjusts the weights of loss components, such as mean squared error (MSE) and Kullback-Leibler Divergence (KLD), to optimize tasks like autoencoders and variational autoencoders. Similarly, the dynamically weighted balanced (DWB) loss proposed by Ruwani et al. \citep{fernando2021dynamically} improves accuracy in class-imbalanced datasets by adjusting weights based on class frequency and prediction difficulty. These approaches highlight the importance of adaptability in achieving balanced learning.

Dynamic weighting has also proven effective in specialized applications. In unsupervised image segmentation, Guermazi et al. \citep{guermazi2022dynamically} developed a dynamically weighted loss function that balances feature similarity and spatial continuity during training. Song et al. \citep{song2023dynamic} introduced a similar approach for multiorgan segmentation, where the weight assigned to each organ is adjusted based on its learning difficulty, ensuring consistent performance across organs with varying sizes and complexities.

Beyond segmentation, adaptive loss functions have been applied to tasks such as image fusion and classification. FuseFormer \citep{erdogan2024fuseformer} and S2F-Net \citep{zhao2024s2f} balance pixel-level accuracy and structural similarity for infrared and visible image fusion. In finger vein recognition, combining cosine softmax and triplet loss improves feature learning and robustness \citep{ou2021fusion}. For ultrasound diagnosis systems, fixed-weight combinations of CAM-based and QUS-based losses enhance lesion discrimination \citep{tasnim2023cam}. Meta-learning techniques, such as those proposed by Jiang et al. \citep{jiang2023dynamic}, further demonstrate the potential of dynamic loss weighting to adapt to label noise and class imbalance in biased datasets.

Building on these advancements, the DMF framework introduces a novel approach to multi-loss function optimization. Unlike fixed-weight strategies, DMF leverages historical loss data to dynamically adjust the weighting of individual loss functions in real time. This ensures a balanced focus on different aspects of the learning task, preventing the model from becoming overly biased toward any single component. Our framework draws inspiration from prior work on composite loss functions and dynamic weighting but extends these concepts to provide a more adaptive and scalable solution for deep learning applications.

The contributions of this paper address fundamental challenges in medical image segmentation, including class imbalance and the manual tuning of multiple loss functions. Specifically, our key contributions are as follows:

\textbf{Dynamic Memory Fusion Framework}: We propose a novel framework for adaptive multi-loss function optimization, enabling real-time adjustment of loss weights based on historical data.
\textbf{Class-Balanced Loss Function}: We introduce a new loss function designed to address class imbalance, prioritizing underrepresented classes and improving segmentation performance for minority classes.
\textbf{Comprehensive Evaluation}: We conduct experiments on three breast ultrasound datasets (BUSI, BUSC, and BUS Synthetic), demonstrating the effectiveness of the DMF framework across various metrics.
\textbf{Open-Source Code}: To ensure reproducibility and encourage further research, we provide an open-source implementation of the DMF framework.
By addressing the limitations of existing approaches and introducing innovations in dynamic weighting and class balance, this work aims to advance the state of the art in medical image segmentation and beyond.

\section{Background}

Deep learning tasks, especially in the domains of segmentation, classification, and regression, have increasingly relied on composite and dynamically weighted loss functions to address challenges such as class imbalance, boundary precision, and feature learning. A growing body of research has explored how combining multiple loss functions and dynamically adjusting their weights can lead to more effective optimization, particularly for tasks requiring complex decisions at multiple scales and resolutions.

In medical imaging, composite loss functions have become popular in improving segmentation accuracy. Jesson et al. \citep{jesson2017brain} introduced a novel approach to brain tumor segmentation using a 3D fully connected network (FCN) with a multi-scale loss function. This composite loss ensures learning at different spatial resolutions by comparing predictions with down-sampled ground truth, addressing class imbalance through a curriculum sample weighting strategy. Similarly, Sharifzadeh et al. \citep{sharifzadeh2023phase} developed an adaptive mixed loss function for ultrasound imaging, gradually shifting from more straightforward B-mode data to more complex RF data. This progressive approach improves reconstruction accuracy and avoids local minima during training, resembling curriculum learning techniques.

Dynamic and composite loss functions have also been employed in diverse fields beyond medical imaging. For example, Lv et al. \citep{lv2024ssdfusion} introduced SSDFusion, which leverages a composite loss combining fusion and segmentation losses to fuse infrared and visible images effectively while maintaining semantic accuracy. Similarly, Gao et al. \citep{gao2019multi} enhanced face recognition performance by fusing multiple loss functions automatically, learning the optimal balance between intra-class compactness and inter-class separation. Moreover, attention-based models, such as the attention-gate medical transformer (AGMT) for ultrasound image segmentation, incorporate specialized loss functions like the average radial derivative increment ($\Delta$ARD) to improve shape feature detection, while other methods like $\delta$ARD further enhance the sensitivity to contours and regions in low-contrast medical images \citep{zhao2024learning, zhao2024loss}.

Adaptive loss functions have proven effective in segmentation tasks where visual detail is crucial. Chao et al. \citep{chao2020multi} introduced an adaptive composite loss function for a multi-field-of-view (FoV) visual saliency model in 360° image processing. This method dynamically adjusts the weights of KLD, normalized scanpath saliency (NSS), and linear correlation coefficient (CC) based on their standard deviations during training. By doing so, the model optimizes performance by balancing key saliency features. Similarly, B et al. \citep{sushma2024aapfc} developed a deep learning model for breast lesion segmentation in ultrasound images using a composite loss that combines binary cross-entropy (BCE) and boundary loss, demonstrating the efficacy of composite loss functions in medical image analysis.

Dynamic loss weighting methods have also been introduced to tackle imbalanced datasets and optimize neural network training. Mang et al. \citep{mang2024dynamic} presented DYNAWEIL, a dynamic weighted loss function for solving partial differential equations (PDEs) on imbalanced data. By adjusting the weights of the loss based on historical errors during training, DYNAWEIL improves prediction accuracy in regions with high errors. Similarly, Lu et al. \citep{lu2019dynamic} introduced dynamic weighted cross entropy (DWCE) for semantic segmentation, where cross-entropy loss is dynamically weighted based on class proportion to handle class imbalance effectively. Additionally, Maldonado et al. \citep{maldonado2023owadapt} introduce an adaptive loss function based on ordered weighted averaging (OWA) operators, dynamically adjusting class weights to handle class imbalance and noise, improving classification performance over traditional methods like focal loss and cross-entropy.

Dynamic loss weighting also finds applications in sentiment classification, as demonstrated by Runda et al. \citep{runda2020lstm}, who introduced a model that adjusts the weights of loss functions based on class prediction error rates, ensuring stable training even in imbalanced datasets. A problem also addressed by Roy et al. \citep{roy2023margin} in their margin-aware adaptive-weighted loss (MAAW), which enhances intraclass compactness and interclass separability by dynamically adjusting weights for hard-to-train samples using confidence scores.

In tasks requiring the balance of multiple cost functions, novel dynamic loss weighting strategies have shown significant performance improvements. Groenendijk et al. \citep{groenendijk2021multi} introduced CoV-weighting, a multi-loss weighting strategy for single-task learning problems such as depth estimation and semantic segmentation. By utilizing the coefficient of variation, CoV-weighting dynamically adjusts the weights of loss functions, eliminating the need for manual weight tuning and grid searches. Ocampo et al. \citep{ocampo2024adaptive} extended this idea by introducing an adaptive loss weighting approach for machine learning interatomic potentials (ML-IAPs). Using the SoftAdapt algorithm, this method recalibrates the contribution of each loss component based on their loss values, ensuring that no single component dominates the training process. 

Xiang et al. \citep{xiang2022self} introduced lbPINNs, a self-adaptive loss-balanced method for physics-informed neural networks (PINNs). This model adaptively assigns weights to multiple loss terms during training, optimizing them at each epoch through Gaussian probabilistic models. Barron \citep{barron2019general} further extended this concept by proposing a generalized adaptive robust loss function that unifies several well-known loss functions, such as Cauchy and Charbonnier, enabling automatic adjustment during training for improved performance across a range of computer vision tasks.

These studies underscore the importance of combining loss functions and dynamically adjusting their weights to address diverse challenges across deep learning applications. Whether handling class imbalance, ensuring boundary precision, or improving segmentation accuracy, dynamic and composite loss functions provide robust solutions for optimizing neural networks in various tasks.

\section{Methodology}

Manually selecting and weighting multiple loss functions in deep learning can be a complex task, as the relative importance of each cost function may evolve throughout the training process. A fixed weight assignment is often inadequate, as it fails to account for the dynamic nature of learning, where different components may require varying levels of focus at various stages. A more logical approach is to implement a system that can dynamically adjust the weights of each loss function based on their real-time contributions to the model’s overall performance. This framework adjusts weights during the training process in real-time, running in parallel with model training. The computational overhead is minimal, requiring only the storage of a subset of historical loss function values in small arrays. Consequently, memory consumption remains low, ensuring computational efficiency even for complex tasks. This adaptive weighting maintains a balanced and responsive training process, leading to more effective optimization.

We introduce the DMF framework for adaptive loss optimization to address the challenges inherent in deep learning tasks involving multiple loss functions. This section details the loss function's components, strategies for weight adjustment, the integration of auxiliary losses, and the computational considerations involved.

\subsection{Dynamic Memory Fusion Framework}

The primary objective of the proposed framework is to minimize a DMF loss function, denoted as $\mathcal{L}$, which integrates multiple base loss functions with an auxiliary loss that addresses specific task-related challenges, such as improving generalization or mitigating bias. The total loss function is defined as:

\begin{equation} 
\mathcal{L}(y, \hat{y}, t) = \sum_{i=1}^{N} w_i \mathcal{L}_i(y, \hat{y}) + \gamma(t) \mathcal{L}_{a}(y, \hat{y}) 
\end{equation}

\noindent where $y$ and $\hat{y}$ refer to ground truth and predicted labels, respectively; $t$ refers to the cycle completed (which could be a batch or epoch), $N$ represents the number of base loss functions, $\mathcal{L}_i$ are the individual loss functions, $w_i$ are adaptive weights dynamically adjust based on the historical data of each loss function, $\mathcal{L}_{a}$ denotes the auxiliary loss, which can be any task-specific function critical in early training stages, $\gamma$ is a scaling factor controlling the contribution of the auxiliary loss, and $t$ indicates the current training step.

This formulation allows the model to adaptively prioritize different loss functions throughout the training process, ensuring a balanced and optimized performance while also incorporating essential task-specific considerations through the auxiliary loss function.

\subsection{Regularizing Loss Function}

The auxiliary loss function, denoted as $\mathcal{L}_{a}$, is integral to guiding the model during the early stages of training by incorporating additional objectives crucial for the learning process. The definition of the auxiliary loss is flexible and can be tailored to the specific task and objectives at hand. The auxiliary loss is typically designed to emphasize aspects of the model that require special attention during training, particularly in the early phases when the model is still learning foundational patterns in the data. To ensure that the influence of the auxiliary loss is most potent during these initial stages and diminishes as training progresses, it is scaled by an exponential decay factor $\gamma$, which is defined as:

\begin{equation} 
\gamma(t) = \gamma_0 e^{-\tau t}
\end{equation}

\noindent where $\gamma_0$ represents the initial weight of the auxiliary loss, and $\tau$ is the decay rate. This scaling mechanism allows the auxiliary loss to influence the model during the early training steps, guiding the learning process toward specific goals. As the training progresses and the model converges, the primary loss function gradually takes precedence, allowing for a balanced and comprehensive optimization process.

\section{Dynamic Weight Adjustment with Memory}

The proposed framework's dynamic adjustment of weights $w_i$ for each loss function is driven by historical loss-value data, referred to as memory. This section details the role of memory in the weight adjustment process and its impact on optimization, as well as the mechanism and strategies of the DMF for weight adjustment.

\subsection{Memory Mechanism}
In the DMF framework, memory plays a pivotal role in adapting the weights of the loss functions over time. Each loss function maintains a record of its historical loss values, which are utilized to compute statistical measures that guide the adjustment of weights. The DMF framework employs three memory-based weighting strategies: variance-based weighting, median absolute deviation (MAD)-based weighting, and Bayesian-based weighting.

\subsection{Variance-Based Weighting}

The variance-based weighting approach adjusts the weight $w_i$ of each loss function according to the variance of its loss history $\mathcal{H}_i$. The variance is calculated as follows:

\begin{equation}
Var(\mathcal{H}_i) = \frac{1}{|\mathcal{H}_i|} \sum_{k=1}^{|\mathcal{H}_i|} \left( \mathcal{H}_i^{(k)} - \bar{\mathcal{H}}_i \right)^2,
\end{equation}

\noindent where $|\mathcal{H}_i|$ is the size of the loss history for the $i$-th loss function, $\mathcal{H}_i^{(k)}$ denotes the $k$-th value in the loss history, and $\bar{\mathcal{H}}_i$ represents the mean of the loss history. 

The weight $w_i$ with $N$ base loss functions is then normalized to ensure a balanced distribution:

\begin{equation}
w_i = \frac{Var(\mathcal{H}_i)}{\sum_{j=1}^{N} Var(\mathcal{H}_j)},
\end{equation}

This normalization ensures that loss functions exhibiting higher variance are assigned proportionally greater weight within the interval $[0, 1]$, reflecting their increased contribution to the overall optimization process.

\subsection{MAD-Based Weighting}

The MAD-based weighting approach offers robustness against outliers by utilizing the median absolute deviation from the median of the loss history. The MAD for the $i$-th loss function is computed as:

\begin{equation} 
MAD(\mathcal{H}_i) = median\left( \left| \mathcal{H}_i^{(k)} - median(\mathcal{H}_i) \right| \right)
\end{equation}

\noindent where $median(\cdot)$ represents the median operation. The weight $w_i$ for $N$ base loss functions is then determined by:

\begin{equation} 
w_i = \frac{(MAD(\mathcal{H}_i))^{-1}}{\sum_{j=1}^{N} (MAD(\mathcal{H}_j))^{-1}}
\end{equation}

This approach favors loss functions with lower deviations from their median values, ensuring that more stable loss functions are weighted more heavily.

\subsection{Bayesian-Based Weighting}

The Bayesian-based weighting strategy updates weights based on likelihoods derived from the MAD values. The likelihood for the $i$-th loss function is given by:

\begin{equation}
p(MAD_i | data) \propto \frac{1}{MAD(\mathcal{H}_i)}
\end{equation}

The posterior distribution for the weight $w_i$ for $N$ base loss functions is then calculated as:

\begin{equation}
w_i = \frac{p(MAD_i | data) \cdot p_i}{\sum_{j=1}^{N} p(MAD_j | data) \cdot p_j}
\end{equation}

\noindent where $p_i$ represents the prior probability for the $i$-th loss function. This approach integrates prior knowledge with observed data, allowing for a more informed and adaptive weighting process.

\subsection{Comparison of Weighting Methods}

The weighting methods offer distinct approaches to dynamically adjusting the contributions of loss functions during model training. The variance-based weighting method emphasizes loss functions exhibiting higher variability in performance under the assumption that increased variance indicates a more dynamic or complex contribution to the optimization process. Loss functions with greater variability are considered to encapsulate more challenging aspects of the task, thus requiring additional attention during training. By adaptively increasing the weights of these high-variance functions, the model shifts its focus toward aspects of the task that may otherwise hinder convergence, thereby promoting a more balanced and efficient optimization.

In contrast, the MAD-based weighting method prioritizes stability by assigning higher weights to loss functions with smaller deviations from their median, signifying more consistent performance over time. This method is designed to enhance robustness to outliers, ensuring that stable loss functions are emphasized while mitigating the impact of abrupt fluctuations. By reducing the likelihood of the model overreacting to transient changes, MAD-based weighting supports a more stable and reliable training process, safeguarding against destabilization.

The Bayesian-based weighting method integrates both prior knowledge and real-time data to update loss function weights dynamically. By assessing the likelihood of each loss function's contribution to the overall performance, this method allows the model to combine predefined expectations, such as the relative importance of certain tasks, with observed performance data. This results in a more informed and adaptive training strategy, where the model balances accumulated learning with prior criteria, leading to a more refined and targeted optimization process. Summarized comparative analysis is as follow:

\begin{itemize}
    \item \textbf{Variance-Based Weighting}: Focuses on loss functions with higher variability, assuming they represent challenging aspects of the task.

    \item \textbf{MAD-Based Weighting}: Prioritizes stability by emphasizing loss functions with smaller deviations from their median values, making it robust to outliers.

    \item \textbf{Bayesian-Based Weighting}: Integrates prior knowledge with observed data, enabling informed and adaptive weight adjustments.

\end{itemize}

\subsection{Normalization of Loss Histories}

To ensure that the weighting mechanism is fair and unaffected by the scale of individual loss values, we apply normalization techniques to the loss histories $\mathcal{H}_i$ before computing statistical measures. Two primary normalization methods are used: min-max scaling and symmetric log scaling.

\paragraph{Min-Max Scaling}

Min-max scaling normalizes the loss history $\mathcal{H}_i$ of each loss function to a fixed range, typically between 0 and 1. This scaling is crucial to prevent loss functions with larger absolute values from dominating the weighting process. The min-max scaling is defined as:

\begin{equation}
\mathcal{H}_i^{scaled} = \frac{\mathcal{H}_i - \min(\mathcal{H}_i)}{\max(\mathcal{H}_i) - \min(\mathcal{H}_i)}
\end{equation}

\noindent where $\min(\mathcal{H}_i)$ and $\max(\mathcal{H}_i)$ are the minimum and maximum values of the loss history for the $i$-th loss function. This method ensures that all loss functions contribute equally, regardless of their original scale.

\paragraph{Symmetric Log Scaling}

While min-max scaling addresses the scale issue, it does not handle cases where the distribution of loss values is skewed. Symmetric log scaling is applied to better handle such skewness, particularly when loss values span several orders of magnitude or contain both positive and negative values. The symmetric log scaling function is defined as:

\begin{equation}
\mathcal{H}_i^{log-scaled} = 
\begin{cases}
\log(1 + \mathcal{H}_i), & if   \quad   \mathcal{H}_i \geq 0 \\
-\log(1 - \mathcal{H}_i), & if   \quad   \mathcal{H}_i < 0
\end{cases}
\end{equation}

By compressing the range of loss values, symmetric log scaling minimizes the impact of outliers, ensuring that variance and MAD calculations become more resilient and accurately reflect the overall behavior of the loss function. 

Consequently, we first apply symmetric log normalization to stabilize the data distribution and then use min-max normalization to rescale the values within the range of 0 to 1. By first stabilizing the data with symmetric log scaling and then normalizing the range with min-max scaling, the process ensures that both the distribution and scale of the loss values are handled appropriately, leading to more robust and effective weight adjustments during training.

\subsection{Adaptive Weighting and the Challenges of Unbounded Weighting}

In the DMF framework, the weights assigned to multiple loss functions are typically constrained to sum to 1. This ensures that the contributions of each loss function are proportionally balanced and normalized, preventing any single component from disproportionately affecting the optimization process. This approach is standard in multi-loss function optimization, as it promotes balanced optimization and reduces the risk of overemphasizing any particular loss function.

Without this constraint, however, several challenges can arise. First, unbounded growth in certain weights may occur, allowing specific components to dominate the optimization and potentially leading to the neglect of other vital tasks. This imbalance can distort the learning process, resulting in suboptimal performance across the model's objectives. Additionally, unconstrained weights may introduce instability into the optimization process, as the overemphasis on certain loss functions can cause erratic gradients, oscillating performance, and difficulties in reaching convergence. Delayed convergence is another potential issue, as the model may frequently shift focus between tasks, prolonging the training process and hindering the achievement of a stable solution. These risks emphasize the importance of maintaining the weight sum constraint to ensure stable and efficient training.

\section{Evaluation Metrics}

To evaluate the performance of our method, we use a range of standard metrics, including dice, intersection over union (IoU), precision, recall, and f1-score. These metrics comprehensively assess the model's ability to segment and classify pixels correctly. The dice score measures the similarity between predicted and ground truth segments, emphasizing precision and recall. IoU evaluates the overlap between the predicted and actual segments, providing insight into how well the prediction captures the object. Precision quantifies how many of the predicted positives are true, while accuracy measures the overall correctness of the predictions, considering both positive and negative classifications. Together, these metrics offer a detailed view of model performance across multiple aspects of segmentation and classification.

\begin{equation}
    Dice = \frac{2 \times TP}{2 \times TP + FP + FN}
\end{equation}

\begin{equation}
    IoU = \frac{TP}{TP + FP + FN}
\end{equation}

\begin{equation}
    Precision = \frac{TP}{TP + FP}
\end{equation}

\begin{equation}
    Recall = \frac{TP}{TP + FN}
\end{equation}

\begin{equation}
    \textit{F1-score} = \frac{2 \times Precision \times Recall}{Precision + Recall}
\end{equation}

\noindent where $TP$ represents the number of true positives, correctly predicted pixels in the positive class; $TN$ is the number of true negatives, correctly predicted pixels in the negative class; $FP$ refers to false positives, or incorrectly predicted positive pixels; and $FN$ denotes false negatives, where pixels that should have been predicted as positive were missed.

\section{Loss Functions}

In semantic segmentation tasks, loss functions are crucial in guiding the optimization process. Different loss functions minimize the error between the predicted segmentation and the ground truth. For our model, we use several loss functions, including categorical cross-entropy, mean intersection over union (mean IoU), mean dice loss, Focal loss, and Tversky loss, each designed to optimize different aspects of segmentation performance.

\subsection{Categorical cross-entropy}

Categorical cross-entropy is a widely used loss function in multi-class segmentation tasks. It measures the dissimilarity between the predicted class probabilities and the true class distribution. For a set of $C$ classes, the categorical cross-entropy loss is defined as:

\begin{equation}
\mathcal{L}_{CE} = -\frac{1}{N} \sum_{i=1}^{N} \sum_{c=1}^{C} y_{ic} \log(\hat{y}_{ic})
\end{equation}

\noindent where $N$ is the number of pixels, $C$ is the number of classes, $y_{ic}$ is the ground truth probability that pixel $i$ belongs to class $c$, and $\hat{y}_{ic}$ is the predicted probability for pixel $i$ in class $c$.

\subsection{Mean Intersection over Union (Mean IoU)}

Mean IoU is used to evaluate the overlap between the predicted segmentation and the ground truth. It is also a loss function by optimizing the average IoU over all classes. Mean IoU is computed as:

\begin{equation}
\mathcal{L}_{IoU} = 1 - \frac{1}{N_c} \sum_{c=1}^{N_c} \frac{TP_c}{TP_c + FP_c + FN_c}
\end{equation}

\noindent where $c$ represents the class and $N_c$ is the number of classes.

\subsection{Mean Dice Loss}

The mean dice loss is derived from the dice similarity coefficient and is widely used to measure overlap between two sets. For segmentation, it can be adapted as a loss function to optimize the dice score for all classes:

\begin{equation}
\mathcal{L}_{Dice} = 1 - \frac{1}{N_c} \sum_{c=1}^{N_c} \frac{2 \times TP_c}{2 \times TP_c + FP_c + FN_c}
\end{equation}

\subsection{Focal Loss}

Focal loss addresses class imbalance by assigning more weight to hard-to-classify examples, reducing the contribution of easy examples. It modifies the cross-entropy loss with a modulating factor:

\begin{equation}
\mathcal{L}_{Focal} = -\frac{1}{N} \sum_{i=1}^{N} \sum_{c=1}^{C} \alpha_c (1 - \hat{y}_{ic})^\gamma  y_{ic} \log(\hat{y}_{ic})
\end{equation}

\noindent where $\alpha_c$ is a balancing factor for class $c$, $\gamma$, is a focusing parameter that adjusts the rate at which easy examples are down-weighted, $y_{ic}$ is the ground truth, and $\hat{y}_{ic}$ is the predicted probability. The values for the balancing factor ($\alpha_c$ = 0.25) and focusing parameter ($\gamma$ = 2) in Focal Loss were empirically optimized to address class imbalance and down-weight easy examples, respectively, based on prior literature \citep{mrad2021machine} and experimental validation.

\subsection{Tversky Loss}

Tversky loss is a generalization of dice loss that allows for a tunable trade-off between false positives and false negatives. It is advantageous when there is a significant class imbalance. The Tversky loss is defined as:

\begin{equation}
\mathcal{L}_{Tversky} = 1 - \frac{1}{C} \sum_{c=1}^{C} \frac{TP_c}{TP_c + \alpha \cdot FP_c + \beta \cdot FN_c}
\end{equation}

\noindent where $\alpha$ and $\beta$ are hyperparameters that control the penalty for FPs and FNs, respectively. This formulation allows the model to be more sensitive to FP or FN based on the specific task requirements. The Tversky Loss parameters were empirically set to 0.7 for $\alpha$ and 0.3 for $\beta$ to balance their trade-off according to task requirements.

\subsection{Class Balanced Dice Loss}

Class imbalance is a common issue in semantic segmentation tasks. Certain classes (e.g., the background) may dominate the image, resulting in biased predictions. To address this, we propose the class-balanced dice (CB-Dice) loss, an extension of the standard dice loss that incorporates class weights based on the pixel distribution across classes. 

The CB-Dice loss is motivated by the need to give more importance to minority classes, often underrepresented in the data. In our implementation, we first calculate the class weights by determining the proportion of pixels that belong to each class. The intuition behind this is that the higher the proportion of a class in the image, the lower its weight should be avoided by biasing the model towards the majority class. These class weights are computed as follows:

\begin{equation}
\textit{class ratio} = \frac{P_i}{P}
\end{equation}

\begin{equation}
w_i = \frac{(\textit{class ratio})^{-1}}{\sum_{i} {(\textit{class ratio})^{-1}}}
\end{equation}

\noindent where $P_i$ represents the number of pixels belonging to class $i$, and $P$ is the total number of pixels in the image. By subtracting the class ratio from 1 (Equation 25), we ensure that classes with fewer pixels receive higher weights, balancing the impact of majority and minority classes on the final loss value. The core of the CB-Dice loss is computed similarly to the traditional dice score but includes class weights. The dice score for each class is computed using the TP, FP, and FN, as follows:

\begin{equation}
Dice_i = \frac{2 \cdot TP_i}{2 \cdot TP_i + FP_i + FN_i}
\end{equation}

The final CB-Dice score is the weighted sum of the individual dice score:

\begin{equation}
\textit{CB-Dice} = \sum_{i} w_i \cdot Dice_i
\end{equation}

The corresponding CB-Dice loss is defined as:

\begin{equation}
\mathcal{L}_{\textit{CB-Dice}} = 1 - \textit{CB-Dice}
\end{equation}

The CB-Dice loss ensures that the optimization process seeks to maximize the CB-Dice score, thereby improving the segmentation performance across all classes, especially the minority ones. This approach leads to a more balanced learning process and helps to mitigate the effects of class imbalance in segmentation tasks.

\begin{figure}[H]
    \centering
    \renewcommand{\arraystretch}{0.6} 
    
    \includegraphics[width = \linewidth]{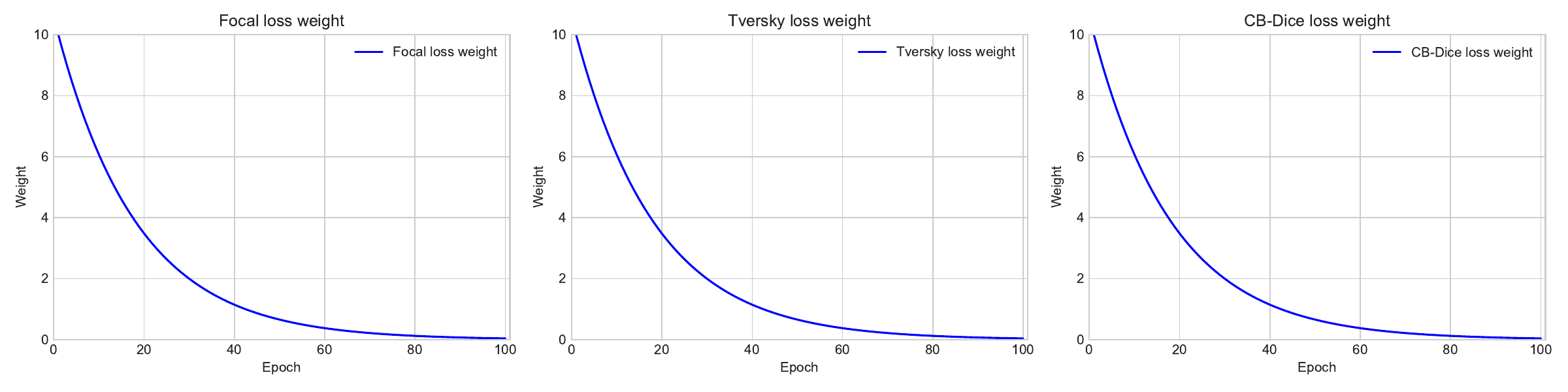} 
    
    \caption{Weight decay of the auxiliary loss using scaling factor ($\gamma_t$) during training.}
    \label{figure:auxiliary_weights}
\end{figure}

The initial weights for base loss functions were set equally to $1/N$, where $N$ represents the number of base loss functions, while the scaling factor ($\gamma_0$) for the auxiliary loss was experimentally determined to be 10, with a decay rate ($\tau$) of 0.05 applied after each batch, ensuring its gradual reduction during training. Figure \ref{figure:auxiliary_weights} shows how the auxiliary loss scaling factor decreases gradually throughout the training process. 

\begin{table}[H]
\centering
\caption{Comparison of Loss Functions}
\label{table:loss_comparison}
\resizebox{\textwidth}{!}{
\begin{tabular}{c|cccc}
\\ [-1.5ex]
Loss Function & Technical Details & Best Use Cases & Strengths & Limitations \\ 
\hline
\\ [-1.5ex]
Categorical Cross-Entropy & Measures label-prediction mismatch & Balanced class tasks & Simple, widely used & Sensitive to imbalance \\ 
\\ [-1.5ex]
\hline
Mean IoU & Evaluates overlap between masks & Segmentation with overlap focus & Focuses on spatial accuracy & Unstable for small classes \\ 
\\ [-1.5ex]
\hline
Mean Dice Loss & Derived from Dice coefficient & Boundary precision tasks & Handles imbalance better & Sensitive to noise \\ 
\\ [-1.5ex]
\hline
Focal Loss & Modulates cross-entropy for hard examples & Imbalanced datasets & Reduces easy example impact & Needs hyperparameter tuning \\ 
\\ [-1.5ex]
\hline
Tversky Loss & Balances false positives/negatives & Tasks needing FP/FN control & Flexible trade-offs & Hyperparameter sensitive \\ 
\\ [-1.5ex]
\hline
CB-Dice Loss & Adds class weights to Dice loss & Severe imbalance tasks & Prioritizes minority classes & Computationally complex \\ 
\\ [-1.5ex]
\end{tabular}
}
\end{table}

\section{Dataset and Materials}

While deep learning classification tasks tend to be less complex and generally involve a single loss function, more challenging tasks often require multiple loss functions to handle fine-grained predictions, boundary accuracy, and class imbalance. This makes them ideal for evaluating adaptive optimization frameworks like DMF. Segmentation tasks, which often involve multiple loss functions, provide a valuable context for testing the DMF framework's effectiveness in addressing the complexities of these tasks. Therefore, our focus on these tasks enables a comprehensive assessment of DMF's capabilities.

This research utilizes breast cancer segmentation as a case study to demonstrate the compatibility and applicability of the DMF framework. By employing the DMF framework, the model can dynamically adjust its focus between multiple cost functions throughout the training process, ensuring that each receives appropriate attention at the right time. To support this investigation, three primary breast ultrasound image segmentation datasets are utilized: the breast ultrasound images dataset, the breast ultrasound cancer dataset, and a third dataset featuring synthetic images generated from a generative model.

\subsection{Breast Ultrasound Segmentation Images}

The Breast Ultrasound Segmentation Images (BUSI) dataset was introduced by Al-Dhabyani et al. \citep{al2020dataset}. It was collected in 2018 at Baheya Hospital for Early Detection \& Treatment of Women's Cancer in Cairo, Egypt. It consists of data from 600 female patients aged 25 to 75 years. The dataset comprises 780 images categorized into normal, benign, and malignant groups. However, to evaluate the segmentation model in this research, only the benign and malignant categories were used, along with their corresponding ground truth masks.

\subsection{Breast Ultrasound Cancer Dataset}

Another dataset used in this research was introduced by Iqbal et al. \citep{iqbal2023unet}, which is an annotated version of the Mendeley dataset \citep{rodrigues2017breast}, referred to as the Breast Ultrasound Cancer (BUSC) dataset. The original Mendeley dataset, consisting of ultrasound images of benign and malignant breast tumors, lacked detailed annotations. However, Iqbal et al. collaborated with experienced radiologists to manually annotate the images, providing ground truth for the segmentation task. The transformed version includes 100 benign and 150 malignant tumor images with annotations; both images and masks have a resolution of 128×128 pixels.

\subsection{Breast Ultrasound Synthetic Dataset}

Additionally, this research incorporated 500 synthetic images of breast tumors created by training a generative model on unannotated breast ultrasound images, referred to as the BUS Synthetic Dataset (BUS), as proposed by Iqbal et al. \citep{iqbal2023unet}. The corresponding probability maps for these synthetic images were generated through a separate model designed to produce segmentation probability outputs. Both the images and masks have a resolution of 128×128 pixels, contributing to a more comprehensive and effective training dataset.

\section{Preprocessing with Bilateral Filtering}

In this research, bilateral filtering was applied during the preprocessing stage to reduce speckle noise in breast ultrasound images. The bilateral filter is particularly useful because it smooths the image while preserving edges, critical for distinguishing between benign and malignant regions in the segmentation task. Studies like \citep{balocco2010srbf} show that speckle-reducing bilateral filtering (SRBF) enhances boundaries while preserving features, making it useful for ultrasound imaging and segmentation. The bilateral filter combines both spatial proximity and pixel intensity in its weighting function. The filtered image $I_{f}(x)$ is computed as:

\begin{equation}
I_{f}(x) = \frac{1}{W_p} \sum_{x_i \in \Omega} I(x_i) \cdot \exp\left(-\frac{|x - x_i|^2}{2\sigma_s^2}\right) \cdot \exp\left(-\frac{|I(x) - I(x_i)|^2}{2\sigma_r^2}\right)
\end{equation}

\noindent where $I(x)$ is the intensity of the pixel at location $x$, $I(x_i)$ is the intensity of the neighboring pixel $x_i$, and $\Omega$ represents the neighborhood around $x$. The parameter $\sigma_s$ controls the spatial proximity weight, determining how much influence nearby pixels have, while $\sigma_r$ controls the intensity difference weight, favoring pixels with similar intensity values. $W_p$ is the normalization factor, which is calculated as:

\begin{equation}
W_p = \sum_{x_i \in \Omega} \exp\left(-\frac{|x - x_i|^2}{2\sigma_s^2}\right) \cdot \exp\left(-\frac{|I(x) - I(x_i)|^2}{2\sigma_r^2}\right)
\end{equation}

This filter effectively reduces speckle noise while preserving the boundaries in ultrasound images.

\section{Results}

\begin{figure}[H]
    \centering 
    \renewcommand{\arraystretch}{0.6} 
    
    \includegraphics[width = 12.5 cm, clip = true, trim = 0pt 350pt 0pt 15pt]{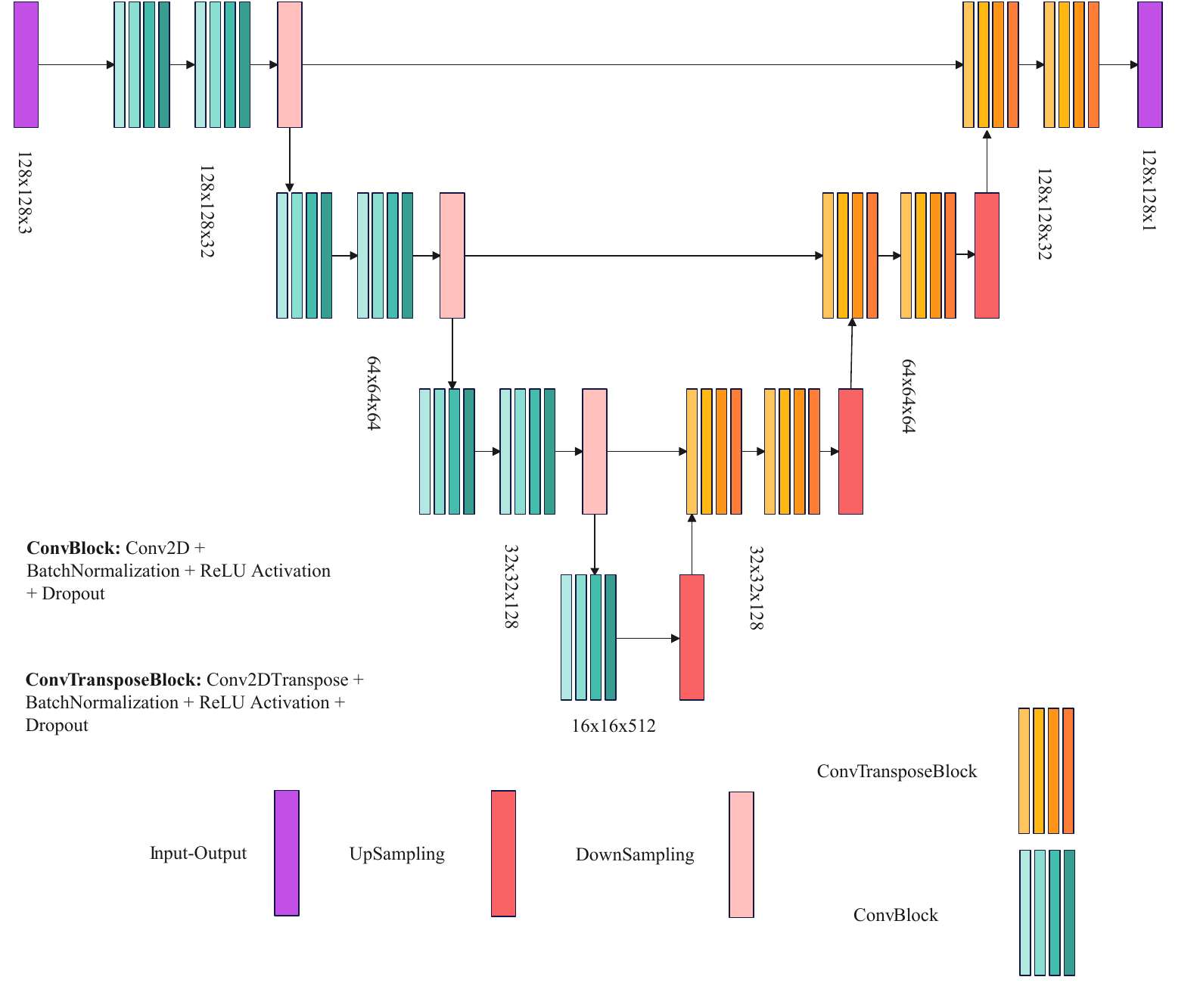} 
    
    \caption{Architecture of the U-Net model used in this study.}
    \label{figure:unet_architecture}
\end{figure}

This section presents the results of applying the DMF framework with different auxiliary loss functions. Each experiment involved training models using three different weighting methods: variance-based, MAD-based, and Bayesian-based. To maintain fairness in comparisons, we utilized a basic U-Net model across all experiments to ensure that the methods, rather than model architectures, are being compared. While more advanced models could enhance performance, our goal is to isolate the benefits of the DMF framework itself. As shown in Figure \ref{figure:unet_architecture}, The U-Net architecture used in this study follows a standard encoder-decoder structure, with convolutional blocks (ConvBlock) for feature extraction and downsampling, followed by transposed convolutional blocks (ConvTransposeBlock) for upsampling and reconstruction. Each block incorporates batch normalization, ReLU activation, and dropout to enhance training stability and generalization, while skip connections between corresponding encoder and decoder layers ensure precise localization of features. 

For every method, ten experiments were conducted with the same weight initialization and data distribution for each experiment, where 70\% of the data was used for training, 15\% for validation, and 15\% for testing, ensuring consistent comparison. In our experiments, we used the Adam optimizer with a learning rate of $1 \times 10^{-3}$ and a batch size of 64, implementing an algorithm to save the optimal weights based on validation dataset, which was originally proposed in our previous work \citep{golnari2024probabilistic}, to prevent overfitting, and trained all models for the same number of epochs without showing signs of overfitting. To address overfitting concerns, in each experiment we applied random shuffling before splitting the dataset into training, validation, and test subsets. This ensured diverse and balanced data exposure across iterations. Table \ref{table:random_shuffling_settings} provides a detailed summary of the experimental configuration.

\begin{table}[H]
\centering
\caption{Experimental Settings Summary – Random Shuffling Method}
\label{table:random_shuffling_settings}
\resizebox{\textwidth}{!}{
\begin{tabular}{l|c}
\\ [-1.5ex]
Parameter & Value / Description \\
\hline

Experiments per dataset & 10 times \\ \hline
Weight initialization & Identical initial weights across each experiment \\ \hline
Data splitting strategy & Random shuffling followed by stratified split \\ \hline
Train/Val/Test ratios & 70\%/15\%/15\% \\ \hline
Augmentation & Enabled with configurable probabilities (real-time) \\ \hline
Batch size & 64 \\ \hline
Learning rate & $1 \times 10^{-3}$ \\ \hline
Epochs per run & Fixed number of epochs (set to 100) \\ \hline
Evaluation metrics & Dice, IoU, F1-Score, Precision, Recall, CB-Dice

\end{tabular}
}
\end{table}

Additionally, regularization techniques (batch normalization, 25\% dropout, and L1L2 regularization), and data augmentation methods such as random flipping, rotations, and shifts used for enhancing model generalization and reducing the risk of overfitting during training. Instead of early stopping, as mentioned above, an algorithm tracked optimal weights based on validation dataset, while learning curves confirmed no divergence between training and validation losses. The evaluation metrics include dice, IoU, f1-score, precision, recall, and CB-Dice. The loss functions used in each experiment comprised categorical cross-entropy, mean IoU, and mean Dice as primary loss functions; additionally, Tversky, Focal, and CB-Dice were utilized as auxiliary loss functions. 

To improve training stability, particularly in the presence of class imbalance, we combined the Mean Dice Loss and Mean IoU Loss as complementary objectives. While these metrics are mathematically related, their gradient behavior during optimization differs, offering a more balanced learning signal that reduces oscillations and improves convergence. This dual-loss strategy enhances robustness by simultaneously encouraging high overlap between predicted and ground truth regions (via Dice) and penalizing spatial mismatches more strictly (via IoU).

The demo source code for implementing the proposed methodology is openly available on GitHub at \href{https://github.com/amingolnari/Demo-Dynamic-Memory-Fusion-Framework}{https://github.com/amingolnari/Demo-Dynamic-Memory-Fusion-Framework}. To facilitate usage, readers can run the code directly in Google Colab. 

Table \ref{table:datasetbusi} compares model performance, reported in percentages, with and without the DMF framework, and various loss functions on the BUSI dataset. Similarly, Table \ref{table:datasetbusc} compares model performance on the BUSC dataset using the same reporting style. Additionally, Table \ref{table:datasetbus} compares model performance on the BUS dataset.


\begin{table}[H]
\centering
\caption{Performance metrics summary using the DMF framework with different auxiliary loss functions, alongside a baseline model without DMF. All models were evaluated on the BUSI dataset.}
\label{table:datasetbusi}
\resizebox{\textwidth}{!}{
\begin{tabular}{c|ccccccc}
\\ [-1.5ex]
Loss Function & Weighting Method & Dice & IoU & F1-score & Precision & Recall & CB-Dice \\ 
\hline
\\ [-1.5ex]

\multirow{3}{*}{Tversky Loss} 
& Variance & 93.30 $\pm$0.91 & 87.91 $\pm$1.44 & 93.36 $\pm$0.92 & 94.37 $\pm$1.02 & 92.28 $\pm$1.46 & 89.36 $\pm$1.76 \\ 
& MAD & 93.75 $\pm$0.69 & 88.64 $\pm$1.12 & 93.82 $\pm$0.68 & 95.31 $\pm$0.70 & 92.38 $\pm$1.07 & 90.17 $\pm$1.37 \\ 
& Bayesian & 90.53 $\pm$1.56 & 83.57 $\pm$2.31 & 90.67 $\pm$1.73 & 92.17 $\pm$3.82 & 89.38 $\pm$0.76 & 85.01 $\pm$3.04 \\ 
\hline
\\ [-1.5ex]

\multirow{3}{*}{Focal Loss} 
& Variance & 94.35 $\pm$1.01 & 89.67 $\pm$1.70 & 94.43 $\pm$0.97 & \textbf{96.12 $\pm$0.61} & 92.84 $\pm$1.71 & 91.13 $\pm$1.68 \\ 
& MAD & 92.67 $\pm$1.02 & 86.88 $\pm$1.63 & 92.98 $\pm$0.91 & 96.95 $\pm$0.37 & 89.41 $\pm$1.52 & 88.41 $\pm$1.72 \\ 
& Bayesian & 93.19 $\pm$1.17 & 87.75 $\pm$1.86 & 93.22 $\pm$1.18 & 94.13 $\pm$1.12 & 92.42 $\pm$1.44 & 89.25 $\pm$2.27 \\ 
\hline
\\ [-1.5ex]

\multirow{3}{*}{CB-Dice Loss} 
& Variance & \textbf{95.22 $\pm$0.53} & \textbf{91.34 $\pm$0.43} & \textbf{95.25 $\pm$0.55} & 95.58 $\pm$0.89 & \textbf{94.93 $\pm$0.87} & \textbf{92.55 $\pm$0.97} \\ 
& MAD & 95.02 $\pm$0.52 & 90.75 $\pm$0.49 & 94.98 $\pm$0.49 & 95.14 $\pm$0.39 & 94.82 $\pm$0.51 & 91.98 $\pm$0.76 \\ 
& Bayesian & 92.60 $\pm$0.46 & 87.42 $\pm$0.30 & 93.31 $\pm$0.53 & 91.56 $\pm$0.18 & 95.12 $\pm$0.87 & 88.94 $\pm$0.32 \\ 
\hline
\\ [-1.5ex]

\multirow{3}{*}{No Auxiliary Loss} 
& Variance & 93.93 $\pm$1.08 & 88.99 $\pm$1.77 & 93.98 $\pm$1.05 & 95.31 $\pm$0.63 & 92.72 $\pm$1.70 & 90.22 $\pm$1.87 \\ 
& MAD & 93.20 $\pm$0.95 & 87.77 $\pm$1.55 & 93.32 $\pm$0.88 & 95.34 $\pm$1.20 & 91.45 $\pm$1.93 & 89.05 $\pm$1.55 \\ 
& Bayesian & 91.97 $\pm$1.05 & 85.51 $\pm$1.64 & 91.98 $\pm$1.06 & 92.03 $\pm$1.60 & 91.99 $\pm$0.67 & 87.09 $\pm$1.85 \\ 
\hline
\\ [-1.5ex]

\multirow{1}{*}{Using Fixed Weights} 
& Not Used & 91.84 $\pm$0.73 & 85.19 $\pm$0.97 & 91.81 $\pm$0.86 & 92.78 $\pm$0.96 & 90.85 $\pm$0.63 & 87.05 $\pm$1.41 \\ 

\end{tabular}
}
\end{table}

The findings from Table \ref{table:datasetbusi} highlight that using the CB-Dice loss function with the variance-based weighting method consistently produced the best results across most metrics, including dice (95.22 $\pm$0.53\%), IoU (91.34 $\pm$0.43\%), f1-score (95.25 $\pm$0.55\%), and CB-Dice (92.55 $\pm$0.97\%). This combination outperformed other loss functions and weighting methods. Models utilizing the Focal and Tversky losses also showed strong performance, particularly when combined with the variance-based and MAD-based weighting methods. Focal loss achieved the highest precision (96.12 $\pm$0.61\%) among all setups. In contrast, models without auxiliary loss or using fixed weights performed noticeably worse, particularly with lower IoU and CB-Dice scores. 

\begin{figure}[H]
    \centering 
    \renewcommand{\arraystretch}{0.6} 
    \setlength{\tabcolsep}{0.4pt} 
    \begin{tabular}{*{8}{>{\centering\arraybackslash}m{0.11\linewidth}}}
        \includegraphics[width=\linewidth]{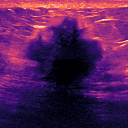}   
      &
        \includegraphics[width=\linewidth]{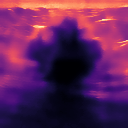}   
      & 
        \includegraphics[width=\linewidth]{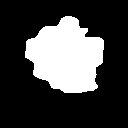}   
      &  
        \includegraphics[width=\linewidth]{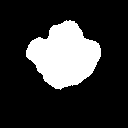}   
      &  
        \includegraphics[width=\linewidth]{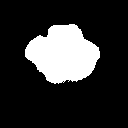}   
      & 
        \includegraphics[width=\linewidth]{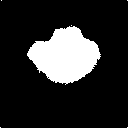}   
      & 
        \includegraphics[width=\linewidth]{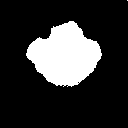}   
      & 
        \includegraphics[width=\linewidth]{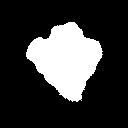}   
     \\

        \includegraphics[width=\linewidth]{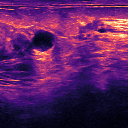}   
      &
        \includegraphics[width=\linewidth]{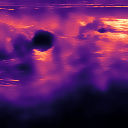}   
      & 
        \includegraphics[width=\linewidth]{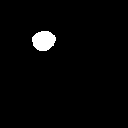}   
      &  
        \includegraphics[width=\linewidth]{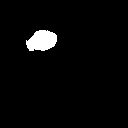}   
      &  
        \includegraphics[width=\linewidth]{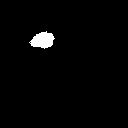}   
      & 
        \includegraphics[width=\linewidth]{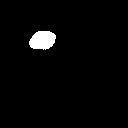}   
      & 
        \includegraphics[width=\linewidth]{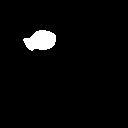}   
      & 
        \includegraphics[width=\linewidth]{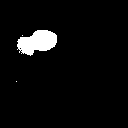}   
     \\

        \includegraphics[width=\linewidth]{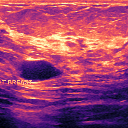}   
      &
        \includegraphics[width=\linewidth]{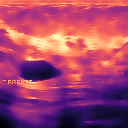}   
      & 
        \includegraphics[width=\linewidth]{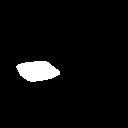}   
      &  
        \includegraphics[width=\linewidth]{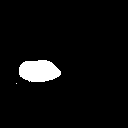}   
      &  
        \includegraphics[width=\linewidth]{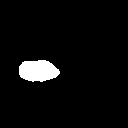}   
      & 
        \includegraphics[width=\linewidth]{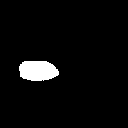}   
      & 
        \includegraphics[width=\linewidth]{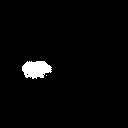}   
      & 
        \includegraphics[width=\linewidth]{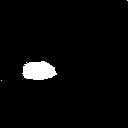}   
     \\

        \includegraphics[width=\linewidth]{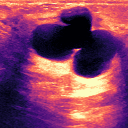}   
      &
        \includegraphics[width=\linewidth]{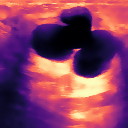}   
      & 
        \includegraphics[width=\linewidth]{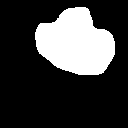}   
      &  
        \includegraphics[width=\linewidth]{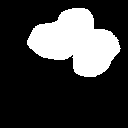}   
      &  
        \includegraphics[width=\linewidth]{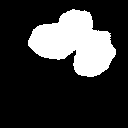}   
      & 
        \includegraphics[width=\linewidth]{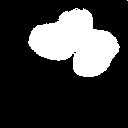}   
      & 
        \includegraphics[width=\linewidth]{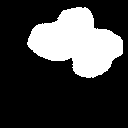}   
      & 
        \includegraphics[width=\linewidth]{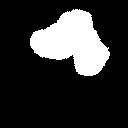}   
     \\

        \includegraphics[width=\linewidth]{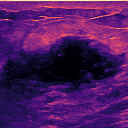}   
      &
        \includegraphics[width=\linewidth]{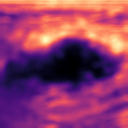}   
      & 
        \includegraphics[width=\linewidth]{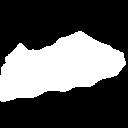}   
      &  
        \includegraphics[width=\linewidth]{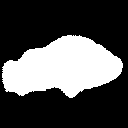}   
      &  
        \includegraphics[width=\linewidth]{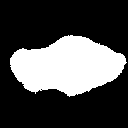}   
      & 
        \includegraphics[width=\linewidth]{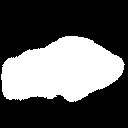}   
      & 
        \includegraphics[width=\linewidth]{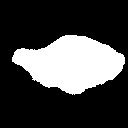}   
      & 
        \includegraphics[width=\linewidth]{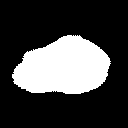}   
     \\

        \includegraphics[width=\linewidth]{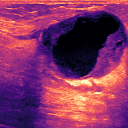}   
      &
        \includegraphics[width=\linewidth]{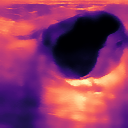}   
      & 
        \includegraphics[width=\linewidth]{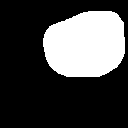}   
      &  
        \includegraphics[width=\linewidth]{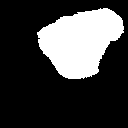}   
      &  
        \includegraphics[width=\linewidth]{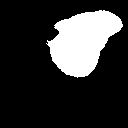}   
      & 
        \includegraphics[width=\linewidth]{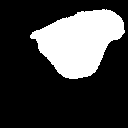}   
      & 
        \includegraphics[width=\linewidth]{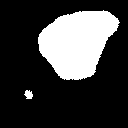}   
      & 
        \includegraphics[width=\linewidth]{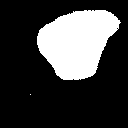}   
     \\

        a & b & c & d & e & f & g & h
    \end{tabular}
    \caption{Each row presents a sample from the BUSI dataset, showing the input breast ultrasound image and the corresponding ground truth mask alongside model predictions. (a) Original breast ultrasound image, (b) Bilaterally filtered image, (c) Ground truth mask, (d) Predicted mask using DMF with Tversky loss as the auxiliary loss function, (e) Predicted mask using DMF with Focal loss as the auxiliary loss function, (f) Predicted mask using DMF with CB-Dice loss as the auxiliary loss function, (g) Predicted mask using DMF without auxiliary loss function, (h) Predicted mask using fixed weights for loss functions.}
    \label{figure:figbusi}
\end{figure}

The comparison of model predictions with different auxiliary loss functions and the ground truth masks is presented in Figure \ref{figure:figbusi}, showcasing the segmentation performance across multiple samples from the BUSI dataset. Each model was selected based on its highest score on the CB-Dice metric.

\begin{table}[H]
\centering
\caption{Performance metrics summary using the DMF framework with different auxiliary loss functions, alongside a baseline model without DMF. All models were evaluated on the BUSC dataset.}
\label{table:datasetbusc}
\resizebox{\textwidth}{!}{
\begin{tabular}{c|ccccccc}
\\ [-1.5ex]
Loss Function & Weighting Method & Dice & IoU & F1-score & Precision & Recall & CB-Dice \\ 
\hline
\\ [-1.5ex]

\multirow{3}{*}{Tversky Loss} 
& Variance & \textbf{95.29 $\pm$0.31} & 91.15 $\pm$0.52 & 95.34 $\pm$0.25 & 94.27 $\pm$0.63 & 96.45 $\pm$0.16 & \textbf{93.68 $\pm$0.44} \\ 
& MAD & 94.79 $\pm$0.69 & 90.35 $\pm$1.13 & 94.85 $\pm$0.65 & 94.36 $\pm$1.31 & 95.41 $\pm$1.59 & 92.93 $\pm$0.90 \\ 
& Bayesian & 94.94 $\pm$0.43 & 90.59 $\pm$0.67 & 94.98 $\pm$0.40 & 94.97 $\pm$0.86 & 95.04 $\pm$1.55 & 92.66 $\pm$1.17 \\ 
\hline
\\ [-1.5ex]

\multirow{3}{*}{Focal Loss} 
& Variance & 94.67 $\pm$0.23 & 90.07 $\pm$0.37 & 94.80 $\pm$0.21 & 92.90 $\pm$0.54 & 96.78 $\pm$0.21 & 92.70 $\pm$0.66 \\ 
& MAD & 94.80 $\pm$0.41 & 90.21 $\pm$0.71 & 94.87 $\pm$0.43 & 93.64 $\pm$0.66 & 96.17 $\pm$1.21 & 92.67 $\pm$0.73 \\ 
& Bayesian & 94.75 $\pm$0.62 & 90.24 $\pm$1.09 & 94.76 $\pm$0.61 & \textbf{95.07 $\pm$0.56} & 94.45 $\pm$0.72 & 92.63 $\pm$0.82 \\ 
\hline
\\ [-1.5ex]

\multirow{3}{*}{CB-Dice Loss} 
& Variance & 95.28 $\pm$0.53 & \textbf{91.75 $\pm$0.32} & \textbf{95.67 $\pm$0.22} & 94.44 $\pm$0.41 & \textbf{96.94 $\pm$0.85} & 93.28 $\pm$0.71 \\ 
& MAD & 94.49 $\pm$0.25 & 89.75 $\pm$0.41 & 94.53 $\pm$0.30 & 94.28 $\pm$0.64 & 94.80 $\pm$1.21 & 92.51 $\pm$0.68 \\ 
& Bayesian & 95.24 $\pm$0.24 & 91.07 $\pm$0.39 & 95.27 $\pm$0.22 & 94.43 $\pm$0.68 & 96.14 $\pm$0.27 & 93.55 $\pm$0.63 \\ 
\hline
\\ [-1.5ex]

\multirow{3}{*}{No Auxiliary Loss} 
& Variance & 94.89 $\pm$0.90 & 89.10 $\pm$1.50 & 94.01 $\pm$0.74 & 94.28 $\pm$0.78 & 93.76 $\pm$1.03 & 92.63 $\pm$0.90 \\ 
& MAD & 93.95 $\pm$1.15 & 89.31 $\pm$0.74 & 93.55 $\pm$0.96 & 93.64 $\pm$1.04 & 93.47 $\pm$1.32 & 92.21 $\pm$1.55 \\ 
& Bayesian & 94.37 $\pm$0.64 & 90.51 $\pm$0.32 & 94.40 $\pm$0.50 & 93.83 $\pm$0.70 & 94.98 $\pm$0.54 & 92.48 $\pm$0.89 \\ 
\hline
\\ [-1.5ex]

\multirow{1}{*}{Using Fixed Weights} 
& Not Used & 93.40 $\pm$0.81 & 89.77 $\pm$1.14 & 94.06 $\pm$0.89 & 93.76 $\pm$1.41 & 94.36 $\pm$0.78 & 92.39 $\pm$1.08 \\ 

\end{tabular}
}
\end{table}

Table \ref{table:datasetbusc} presents that the dice scores range from 93.95\% to 95.29\%, while the IoU values vary between 89.10\% and 91.75\%, reflecting minimal variation between methods. Notably, the CB-Dice loss function with the Variance-based weighting method achieved the highest mean IoU (91.75 $\pm$0.32\%) and nearly the highest dice score (95.28 $\pm$0.53\%), demonstrating superior overall performance. Meanwhile, Tversky loss with variance-based weighting performed similarly well, achieving the highest dice score (95.29 $\pm$0.31\%) and a recall of (96.45 $\pm$0.16\%). The models without auxiliary losses showed slightly lower metrics, with the best dice score (94.89 $\pm$0.90\%) using the variance-based weighting method. These results indicate that, while subtle, the choice of auxiliary loss function and weighting method has only a minor impact on overall model performance in this dataset.

\begin{figure}[H]
    \centering 
    \renewcommand{\arraystretch}{0.6} 
    \setlength{\tabcolsep}{0.4pt} 
    \begin{tabular}{*{8}{>{\centering\arraybackslash}m{0.11\linewidth}}}
        \includegraphics[width=\linewidth]{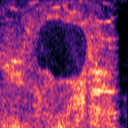}   
      &
        \includegraphics[width=\linewidth]{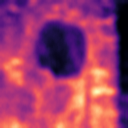}   
      & 
        \includegraphics[width=\linewidth]{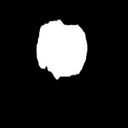}   
      &  
        \includegraphics[width=\linewidth]{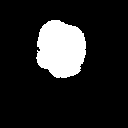}   
      &  
        \includegraphics[width=\linewidth]{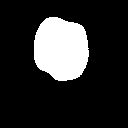}   
      & 
        \includegraphics[width=\linewidth]{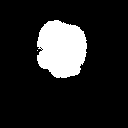}   
      & 
        \includegraphics[width=\linewidth]{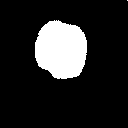}   
      & 
        \includegraphics[width=\linewidth]{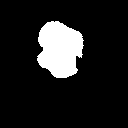}   
     \\

        \includegraphics[width=\linewidth]{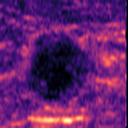}   
      &
        \includegraphics[width=\linewidth]{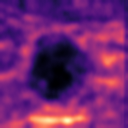}   
      & 
        \includegraphics[width=\linewidth]{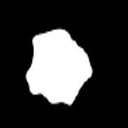}   
      &  
        \includegraphics[width=\linewidth]{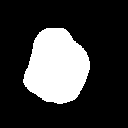}   
      &  
        \includegraphics[width=\linewidth]{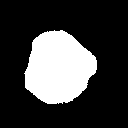}   
      & 
        \includegraphics[width=\linewidth]{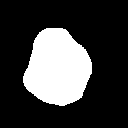}   
      & 
        \includegraphics[width=\linewidth]{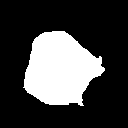}   
      & 
        \includegraphics[width=\linewidth]{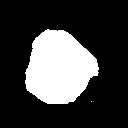}   
     \\

        \includegraphics[width=\linewidth]{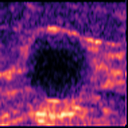}   
      &
        \includegraphics[width=\linewidth]{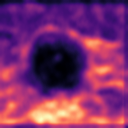}   
      & 
        \includegraphics[width=\linewidth]{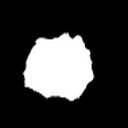}   
      &  
        \includegraphics[width=\linewidth]{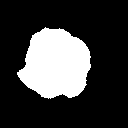}   
      &  
        \includegraphics[width=\linewidth]{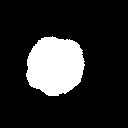}   
      & 
        \includegraphics[width=\linewidth]{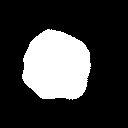}   
      & 
        \includegraphics[width=\linewidth]{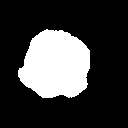}   
      & 
        \includegraphics[width=\linewidth]{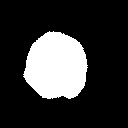}   
     \\

        \includegraphics[width=\linewidth]{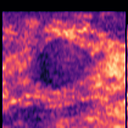}   
      &
        \includegraphics[width=\linewidth]{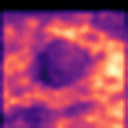}   
      & 
        \includegraphics[width=\linewidth]{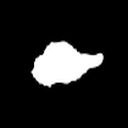}   
      &  
        \includegraphics[width=\linewidth]{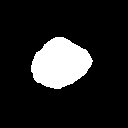}   
      &  
        \includegraphics[width=\linewidth]{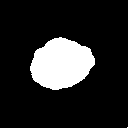}   
      & 
        \includegraphics[width=\linewidth]{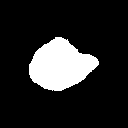}   
      & 
        \includegraphics[width=\linewidth]{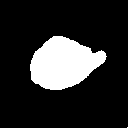}   
      & 
        \includegraphics[width=\linewidth]{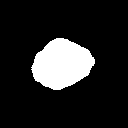}   
     \\

        \includegraphics[width=\linewidth]{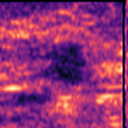}   
      &
        \includegraphics[width=\linewidth]{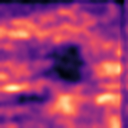}   
      & 
        \includegraphics[width=\linewidth]{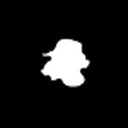}   
      &  
        \includegraphics[width=\linewidth]{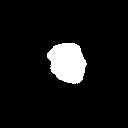}   
      &  
        \includegraphics[width=\linewidth]{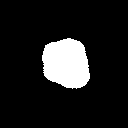}   
      & 
        \includegraphics[width=\linewidth]{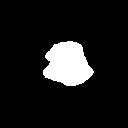}   
      & 
        \includegraphics[width=\linewidth]{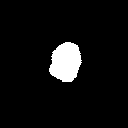}   
      & 
        \includegraphics[width=\linewidth]{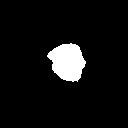}   
     \\

        \includegraphics[width=\linewidth]{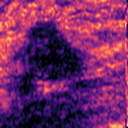}   
      &
        \includegraphics[width=\linewidth]{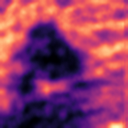}   
      & 
        \includegraphics[width=\linewidth]{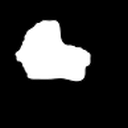}   
      &  
        \includegraphics[width=\linewidth]{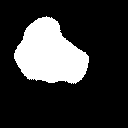}   
      &  
        \includegraphics[width=\linewidth]{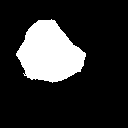}   
      & 
        \includegraphics[width=\linewidth]{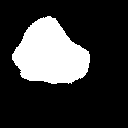}   
      & 
        \includegraphics[width=\linewidth]{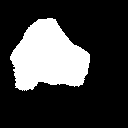}   
      & 
        \includegraphics[width=\linewidth]{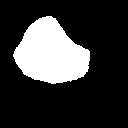}   
     \\

        a & b & c & d & e & f & g & h
    \end{tabular}
    \caption{Each row presents a sample from the BUSC dataset, showing the input breast ultrasound image and the corresponding ground truth mask alongside model predictions. (a) Original breast ultrasound image, (b) Bilaterally filtered image, (c) Ground truth mask, (d) Predicted mask using DMF with Tversky loss as the auxiliary loss function, (e) Predicted mask using DMF with Focal loss as the auxiliary loss function, (f) Predicted mask using DMF with CB-Dice loss as the auxiliary loss function, (g) Predicted mask using DMF without auxiliary loss function, (h) Predicted mask using fixed weights for loss functions.}
    \label{figure:figbusc}
\end{figure}

The comparison of model predictions with different auxiliary loss functions and the ground truth masks is presented in Figure \ref{figure:figbusc}, showcasing the segmentation performance across multiple samples from the BUSC dataset. Each model was selected based on its highest score on the CB-Dice metric.


\begin{table}[H]
\centering
\caption{Performance metrics summary using the DMF framework with different auxiliary loss functions, alongside a baseline model without DMF. All models were evaluated on the BUS dataset.}
\label{table:datasetbus}
\resizebox{\textwidth}{!}{
\begin{tabular}{c|ccccccc}
\\ [-1.5ex]
Loss Function & Weighting Method & Dice & IoU & F1-score & Precision & Recall & CB-Dice \\ 
\hline
\\ [-1.5ex]

\multirow{3}{*}{Tversky Loss} 
& Variance & 93.35 $\pm$0.22 & 91.28 $\pm$0.37 & 95.42 $\pm$0.21 & 97.21 $\pm$0.42 & 93.69 $\pm$0.49 & 93.20 $\pm$0.30 \\ 
& MAD & 92.34 $\pm$0.45 & 86.27 $\pm$0.27 & 92.78 $\pm$0.39 & \textbf{97.33 $\pm$0.27} & 88.86 $\pm$0.64 & 88.72 $\pm$0.65 \\ 
& Bayesian & 94.15 $\pm$0.32 & 89.24 $\pm$0.59 & 94.20 $\pm$0.34 & 95.53 $\pm$0.58 & 93.14 $\pm$0.68 & 91.47 $\pm$0.49 \\ 
\hline
\\ [-1.5ex]

\multirow{3}{*}{Focal Loss} 
& Variance & 94.08 $\pm$0.54 & 89.14 $\pm$0.92 & 94.24 $\pm$0.51 & 96.81 $\pm$0.28 & 91.81 $\pm$0.76 & 91.33 $\pm$0.79 \\ 
& MAD & 93.09 $\pm$0.18 & 87.43 $\pm$0.29 & 93.37 $\pm$0.15 & 96.97 $\pm$0.17 & 90.07 $\pm$0.29 & 89.80 $\pm$0.25 \\ 
& Bayesian & 94.27 $\pm$1.31 & 89.47 $\pm$1.67 & 94.37 $\pm$0.96 & 95.80 $\pm$0.32 & 92.98 $\pm$1.93 & 91.62 $\pm$1.48 \\ 
\hline
\\ [-1.5ex]

\multirow{3}{*}{CB-Dice Loss} 
& Variance & \textbf{95.71 $\pm$0.37} & \textbf{91.90 $\pm$0.63} & \textbf{95.79 $\pm$0.35} & 95.30 $\pm$0.71 & 95.68 $\pm$0.62 & \textbf{93.72 $\pm$0.67} \\ 
& MAD & 95.05 $\pm$0.75 & 91.54 $\pm$0.46 & 95.27 $\pm$0.63 & 93.30 $\pm$0.53 & \textbf{97.31 $\pm$0.84} & 93.65 $\pm$0.38 \\ 
& Bayesian & 92.79 $\pm$0.51 & 87.19 $\pm$0.81 & 93.32 $\pm$0.41 & 90.16 $\pm$0.87 & 96.28 $\pm$0.23 & 89.55 $\pm$0.80 \\ 
\hline
\\ [-1.5ex]

\multirow{3}{*}{No Auxiliary Loss} 
& Variance & 93.99 $\pm$0.36 & 88.96 $\pm$0.62 & 94.11 $\pm$0.35 & 96.24 $\pm$0.30 & 92.06 $\pm$0.48 & 91.20 $\pm$0.55 \\ 
& MAD & 93.26 $\pm$1.58 & 87.81 $\pm$2.58 & 93.57 $\pm$1.34 & 96.94 $\pm$0.12 & 90.47 $\pm$2.58 & 90.08 $\pm$2.40 \\ 
& Bayesian & 93.84 $\pm$0.11 & 88.70 $\pm$0.19 & 93.93 $\pm$0.12 & 95.80 $\pm$0.63 & 92.16 $\pm$0.50 & 90.99 $\pm$0.15 \\ 
\hline
\\ [-1.5ex]

\multirow{1}{*}{Using Fixed Weights} 
& Not Used & 92.16 $\pm$1.38 & 85.34 $\pm$0.78 & 91.29 $\pm$0.67 & 91.64 $\pm$0.89 & 90.93 $\pm$0.76 & 88.28 $\pm$1.46 \\ 

\end{tabular}
}
\end{table}

Notably, the Table \ref{table:datasetbus} highlights that the CB-Dice loss with variance-based weighting outperformed all other configurations, achieving the highest dice score (95.71 $\pm$0.37\%), IoU (91.90 $\pm$0.63\%), f1-score (95.79 $\pm$0.35\%), and CB-Dice (93.72 $\pm$0.67\%), making it the most effective combination. The MAD-based weighting method with CB-Dice loss also showed strong results, particularly in precision (97.31 $\pm$0.84\%). Among the models using Tversky loss, the MAD-based weighting method excelled in precision (97.33 $\pm$0.27\%) and provided a balanced performance across metrics. Meanwhile, models without auxiliary loss had consistently lower results, with the highest dice score of (91.20 $\pm$0.55\%) and IoU of (88.96 $\pm$0.62\%), falling short compared to models with auxiliary losses. 

In summary, Variance-based weighting with CB-Dice loss consistently outperforms other configurations across all datasets, while MAD-based weighting shows strong precision but slightly lower recall on the BUSC dataset, and Bayesian-based weighting performs well on synthetic data but struggles with real-world variability.

\begin{figure}[H]
    \centering 
    \renewcommand{\arraystretch}{0.6} 
    \setlength{\tabcolsep}{0.4pt} 
    \begin{tabular}{*{8}{>{\centering\arraybackslash}m{0.11\linewidth}}}
        \includegraphics[width=\linewidth]{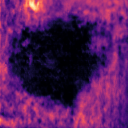}   
      &
        \includegraphics[width=\linewidth]{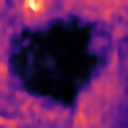}   
      & 
        \includegraphics[width=\linewidth]{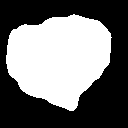}   
      &  
        \includegraphics[width=\linewidth]{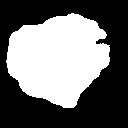}   
      &  
        \includegraphics[width=\linewidth]{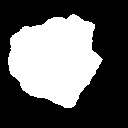}   
      & 
        \includegraphics[width=\linewidth]{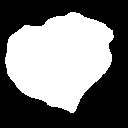}   
      & 
        \includegraphics[width=\linewidth]{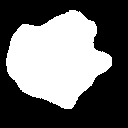}   
      & 
        \includegraphics[width=\linewidth]{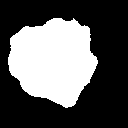}   
     \\

        \includegraphics[width=\linewidth]{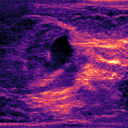}   
      &
        \includegraphics[width=\linewidth]{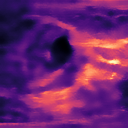}   
      & 
        \includegraphics[width=\linewidth]{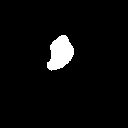}   
      &  
        \includegraphics[width=\linewidth]{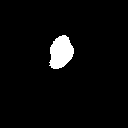}   
      &  
        \includegraphics[width=\linewidth]{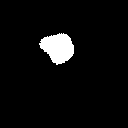}   
      & 
        \includegraphics[width=\linewidth]{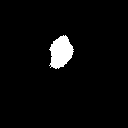}   
      & 
        \includegraphics[width=\linewidth]{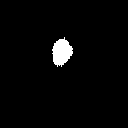}   
      & 
        \includegraphics[width=\linewidth]{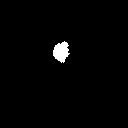}   
     \\

        \includegraphics[width=\linewidth]{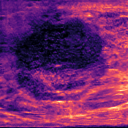}   
      &
        \includegraphics[width=\linewidth]{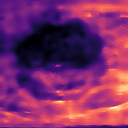}   
      & 
        \includegraphics[width=\linewidth]{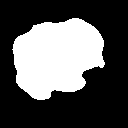}   
      &  
        \includegraphics[width=\linewidth]{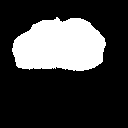}   
      &  
        \includegraphics[width=\linewidth]{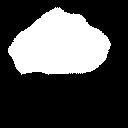}   
      & 
        \includegraphics[width=\linewidth]{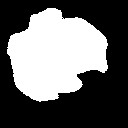}   
      & 
        \includegraphics[width=\linewidth]{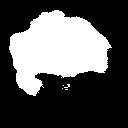}   
      & 
        \includegraphics[width=\linewidth]{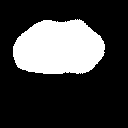}   
     \\

        \includegraphics[width=\linewidth]{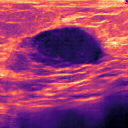}   
      &
        \includegraphics[width=\linewidth]{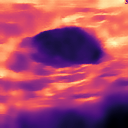}   
      & 
        \includegraphics[width=\linewidth]{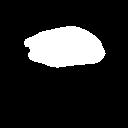}   
      &  
        \includegraphics[width=\linewidth]{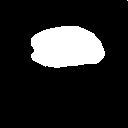}   
      &  
        \includegraphics[width=\linewidth]{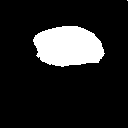}   
      & 
        \includegraphics[width=\linewidth]{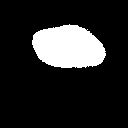}   
      & 
        \includegraphics[width=\linewidth]{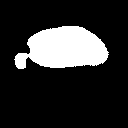}   
      & 
        \includegraphics[width=\linewidth]{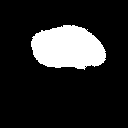}   
     \\

        \includegraphics[width=\linewidth]{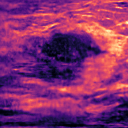}   
      &
        \includegraphics[width=\linewidth]{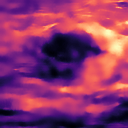}   
      & 
        \includegraphics[width=\linewidth]{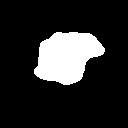}   
      &  
        \includegraphics[width=\linewidth]{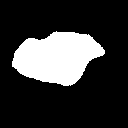}   
      &  
        \includegraphics[width=\linewidth]{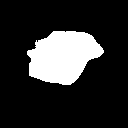}   
      & 
        \includegraphics[width=\linewidth]{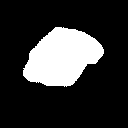}   
      & 
        \includegraphics[width=\linewidth]{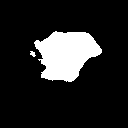}   
      & 
        \includegraphics[width=\linewidth]{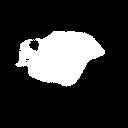}   
     \\

        \includegraphics[width=\linewidth]{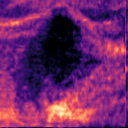}   
      &
        \includegraphics[width=\linewidth]{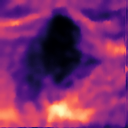}   
      & 
        \includegraphics[width=\linewidth]{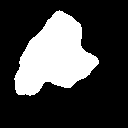}   
      &  
        \includegraphics[width=\linewidth]{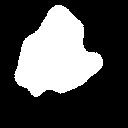}   
      &  
        \includegraphics[width=\linewidth]{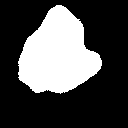}   
      & 
        \includegraphics[width=\linewidth]{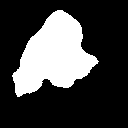}   
      & 
        \includegraphics[width=\linewidth]{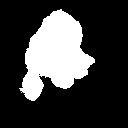}   
      & 
        \includegraphics[width=\linewidth]{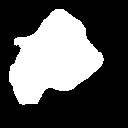}   
     \\

        a & b & c & d & e & f & g & h
    \end{tabular}
    \caption{Each row presents a sample from the BUS dataset, showing the input breast ultrasound image and the corresponding ground truth mask alongside model predictions. (a) Original breast ultrasound image, (b) Bilaterally filtered image, (c) Ground truth mask, (d) Predicted mask using DMF with Tversky loss as the auxiliary loss function, (e) Predicted mask using DMF with Focal loss as the auxiliary loss function, (f) Predicted mask using DMF with CB-Dice loss as the auxiliary loss function, (g) Predicted mask using DMF without auxiliary loss function, (h) Predicted mask using fixed weights for loss functions.}
    \label{figure:figbus}
\end{figure}

The comparison of model predictions with different auxiliary loss functions and the ground truth masks is presented in Figure \ref{figure:figbus}, showcasing the segmentation performance across multiple samples from the BUS dataset. Each model was selected based on its highest score on the CB-Dice metric.

\begin{figure}[H]
    \centering 
    \renewcommand{\arraystretch}{0.6} 
    
    \includegraphics[width = \linewidth]{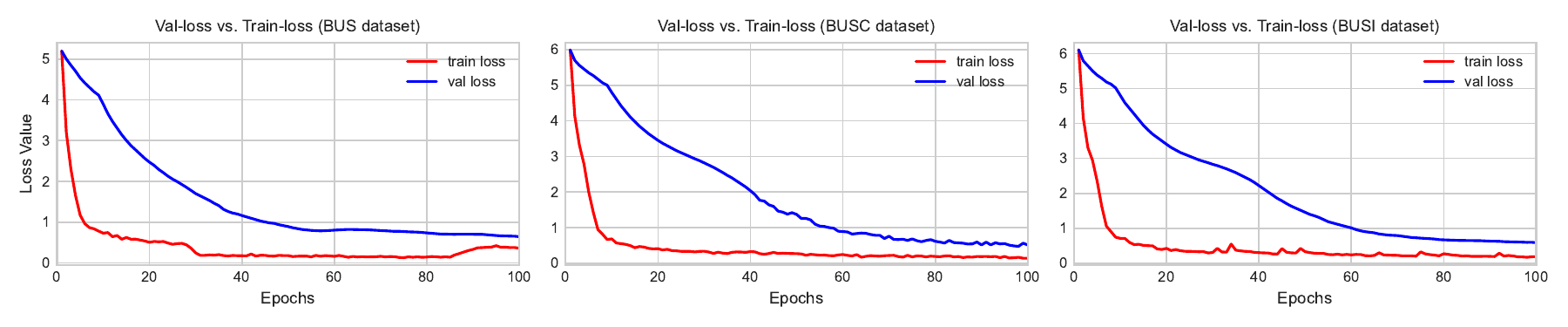} 
    
    \caption{Training and validation loss curves across epochs.}
    \label{figure:val_vs_train}
\end{figure}

Overall, these results demonstrate that the choice of auxiliary loss function and the weighting method play critical roles in optimizing segmentation performance on each dataset. As illustrated in Figure \ref{figure:val_vs_train}, the training and validation loss curves ($\mathcal{L}(y, \hat{y}, t)$) demonstrate a consistent decline without any significant divergence, indicating that the model does not suffer from overfitting during training.

\begin{figure}[H]
    \centering 
    \renewcommand{\arraystretch}{0.8} 
    \setlength{\tabcolsep}{0.5pt} 
    \begin{tabular}{c*{1}{>{\centering\arraybackslash}m{0.11\linewidth}}}
            \includegraphics[width=\linewidth]{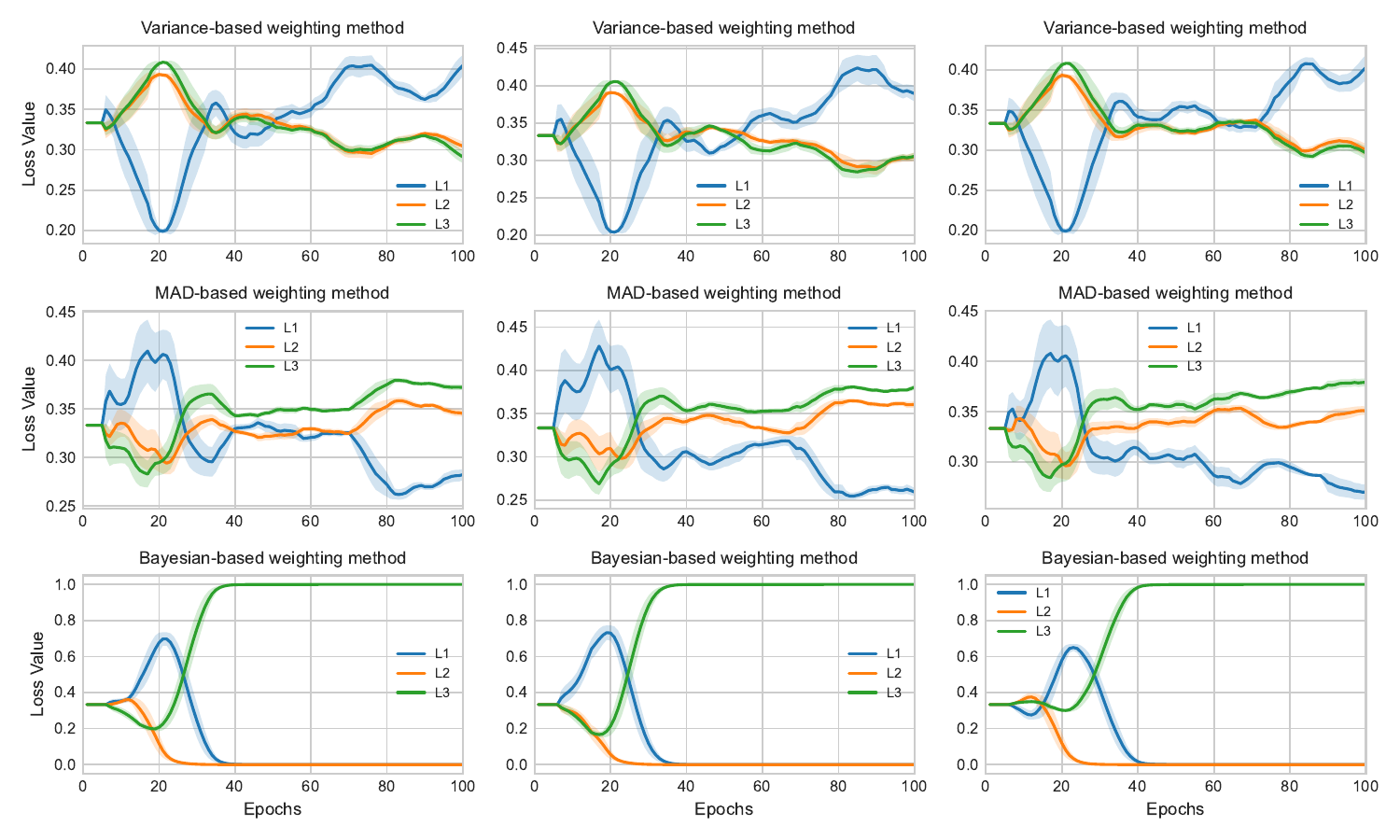}   
\end{tabular}
    \caption{Weight dynamics of W1 in blue for categorical cross-entropy, W2 in red for mean IoU, and W3 in green for mean dice during training in the DMF framework. The first row shows the variance-based method, the second row shows the MAD-based method, and the third row shows the Bayesian-based method. Models are trained on the BUSI dataset}
    \label{figure:figweights}
\end{figure}

Figure  \ref{figure:figweights} shows the dynamic weighting behavior of W1, W2, and W3 during training for categorical cross-entropy, mean IoU, and mean dice loss functions in the DMF framework. It employs three different weighting methods: variance-based, MAD-based, and Bayesian-based. The variance-based method adjusts weights gradually in the early stages, responding to fluctuations in task performance while maintaining a balanced distribution throughout the training process. In contrast, the MAD-based method is more sensitive, resulting in sharper weight adjustments. Although the Bayesian-based method exhibits smooth changes, it adopts a more aggressive approach, focusing on a dominant task during training and maintaining this preference. Each technique offers a unique strategy for handling multi-task loss functions, with different levels of adaptability and task prioritization.

\begin{figure}[H]
    \centering
    \renewcommand{\arraystretch}{0.8}
    \setlength{\tabcolsep}{0.5pt}
    \begin{tabular}{c*{1}{>{\centering\arraybackslash}m{0.11\linewidth}}}
            \includegraphics[width=\linewidth]{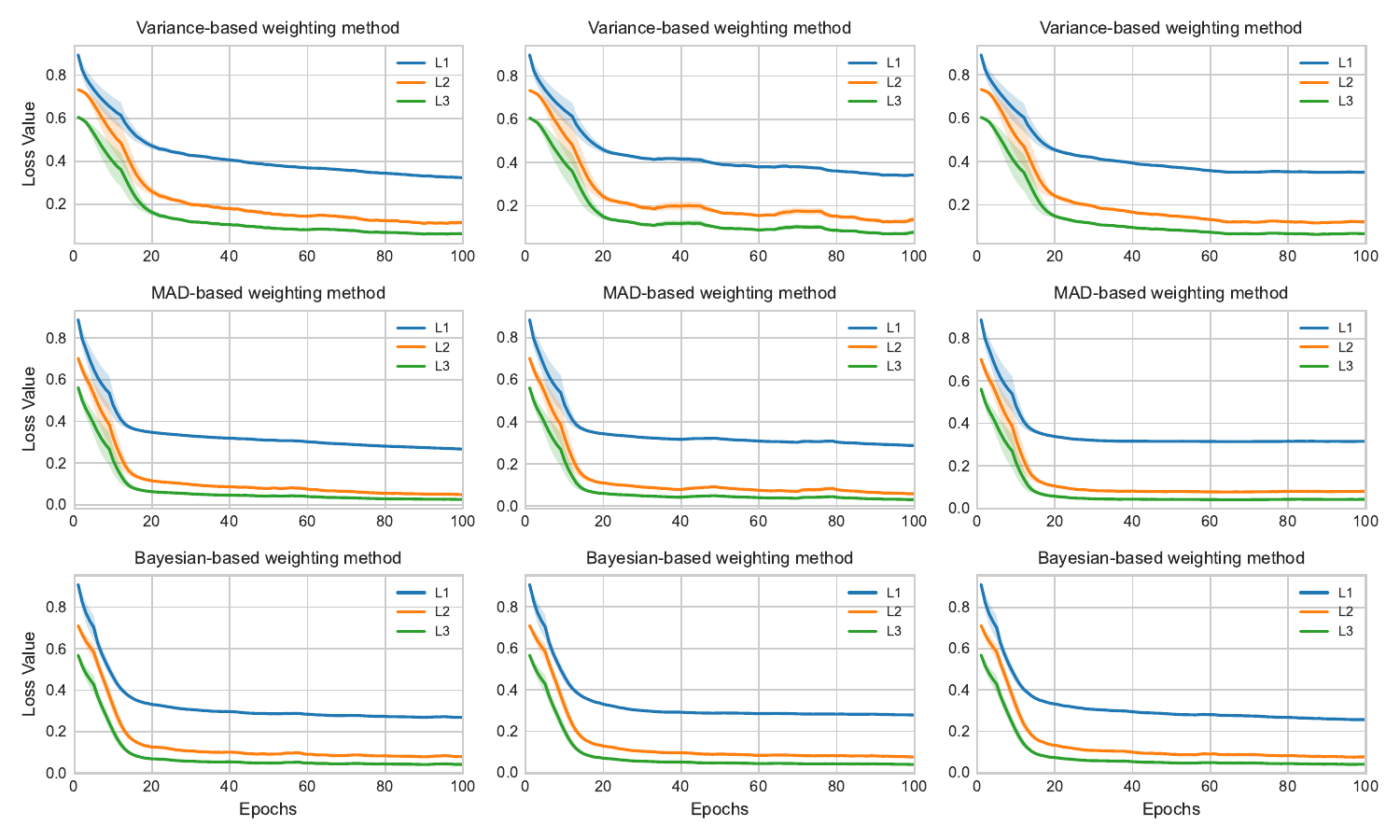}   
\end{tabular}
    \caption{Loss dynamics of L1 in blue for categorical cross-entropy, L2 in red for mean IoU, and L3 in green for mean Dice during training in the DMF framework. The first row shows the variance-based method, the second row shows the MAD-based method, and the third row shows the Bayesian-based method. Models are trained on the BUSI dataset}
    \label{figure:figlosses}
\end{figure}

Figure \ref{figure:figlosses} illustrates the dynamic behavior of the loss functions L1, L2, and L3, respectively, representing categorical cross-entropy, mean IoU, and mean Dice functions within the DMF framework. The figure demonstrates the performance of three different weighting methods: variance-based, MAD-based, and Bayesian-based. The convergence of all loss functions is evident as each steadily decreases without significant fluctuations. 

After a certain number of epochs, there are no considerable changes in the loss values, indicating that they have reached a stable point. This confirms successful optimization convergence, where further training brings minimal improvements. Although the loss values remain stable, the weights continue to adjust (except for the Bayesian-based weighting method, which is aggressive yet smooth in its changes), according to Figure \ref{figure:figweights}. This behavior suggests that the DMF framework avoids local minima and maintains adaptability during training, ensuring robust performance across tasks even in later stages.

To further validate the generalization capability of our method and ensure its reliability across different data distributions, we conducted an additional evaluation using 5-fold cross-validation. In our previous experimental setup, we trained the U-Net 10 times on each dataset with identical initial weights but different train/validation/test splits, achieved through random shuffling. This strategy ensured diverse exposure to the data and provided a stable estimate of model performance. As an extension, we have also applied k-fold cross-validation to further assess the consistency of our results. The fold-averaged performance metrics showed close alignment with those obtained from the repeated random splits, confirming that our findings are not biased by any specific data partitioning strategy and can be reliably generalized. Table \ref{table:kfold_settings} provides a detailed summary of the experimental configuration for the k-fold cross-validation procedure, whereas Table \ref{table:kfold_cv} outlines the main results obtained from the evaluation.

\begin{table}[H]
\centering
\caption{Experimental Settings Summary – k-Fold Cross-Validation}
\label{table:kfold_settings}
\resizebox{\textwidth}{!}{
\begin{tabular}{l|c}
\\ [-1.5ex]
Parameter & Value / Description \\
\hline

Number of folds & 5 folds \\ \hline
Dataset coverage & Each sample used exactly once per fold as test set \\ \hline
Weight initialization & Independent initialization per fold \\ \hline
Data splitting strategy & Stratified k-fold (train/val/test) \\ \hline
Train/Val/Test ratios & 70\%/15\%/15\% \\ \hline
Augmentation & Enabled with configurable probabilities (real-time) \\ \hline
Batch size & 64 \\ \hline
Learning rate & $1 \times 10^{-3}$ \\ \hline
Epochs per run & Fixed number of epochs (set to 100) \\ \hline
Evaluation metrics & Dice, IoU, F1-Score, Precision, Recall, CB-Dice

\end{tabular}
}
\end{table}

\begin{table}[H]
\centering
\caption{Performance metrics obtained through 5-fold cross-validation across all datasets using the CB-Dice loss as the auxiliary loss function. Reported values represent the mean performance across the five folds.}
\label{table:kfold_cv}
\resizebox{\textwidth}{!}{
\begin{tabular}{c|ccccccc}
\\ [-1.5ex]
Dataset & Method & Dice & IoU & F1-score & Precision & Recall & CB-Dice \\ 
\hline
\\ [-1.5ex]

\multirow{3}{*}{BUSI} 
& Variance & 94.98 & 91.57 & 95.25 & 95.89 & 94.62 & 92.83 \\ 
& MAD      & 95.16 & 90.83 & 95.02 & 95.29 & 94.75 & 92.12 \\ 
& Bayesian & 92.73 & 87.48 & 93.28 & 91.37 & 95.27 & 88.87 \\ 
\hline
\\ [-1.5ex]

\multirow{5}{*}{BUSC} 
& Variance & 95.53 & 91.49 & 95.75 & 94.69 & 96.83 & 93.47 \\ 
& MAD      & 94.32 & 89.64 & 94.48 & 94.30 & 94.67 & 92.65 \\ 
& Bayesian & 95.16 & 89.85 & 95.07 & 94.26 & 95.89 & 93.37 \\  
\hline
\\ [-1.5ex]

\multirow{5}{*}{BUS} 
& Variance & 95.84 & 91.69 & 95.47 & 95.48 & 95.46 & 93.87 \\ 
& MAD      & 94.89 & 91.37 & 94.35 & 92.41 & 96.38 & 93.47 \\ 
& Bayesian & 92.98 & 87.26 & 93.60 & 90.65 & 96.73 & 89.76 \\ 

\end{tabular}
}
\end{table}

Table \ref{tab:comp_table} provides a detailed comparison of several existing approaches to dynamic and adaptive loss function weighting, highlighting their respective advantages, limitations, and associated performance outcomes.

\newcolumntype{C}[1]{>{\centering\arraybackslash}p{#1}}

\begin{table}[H]
\centering
\caption{Summary of Dynamic Weighting Methods for Loss Functions}
\label{tab:comp_table}
\resizebox{\textwidth}{!}{ 
\begin{tabular}{C{3cm}|C{5cm}C{4cm}C{4cm}C{4cm}}

Research & Short Description & Pros & Cons & Results \\
\hline
\citep{chao2020multi} & Dynamically balances KLD, CC, and NSS metrics during training based on standard deviations & Adapts to dataset characteristics; improves performance across metrics & Requires tuning of initial weights & Outperforms state-of-the-art methods on multiple datasets \\
\hline
\citep{lu2019dynamic} & Adjusts cross-entropy weights dynamically using batch and global data frequencies & Handles class imbalance effectively; improves segmentation accuracy & Hard truncation may discard useful information & Achieves better performance on imbalanced datasets \\
\hline
\citep{guermazi2022dynamically} & Dynamically adjusts weights between feature similarity and spatial continuity & Balances focus between similarity and continuity; adapts to training dynamics & Requires careful tuning of parameters & Improves semantic segmentation details \\
\hline
\citep{barron2019general} & Dynamically adjusts shape parameter $\alpha$ to transition between traditional loss functions & Eliminates manual tuning; adapts to varying data & May introduce instability if $\alpha$ changes too rapidly & Performs well across diverse tasks and datasets \\
\hline
\citep{ocampo2024adaptive} & Dynamically adjusts weights for energy, forces, and stress during training & Reduces need for manual tuning; balances contributions & Sensitive to initial conditions and learning rates & Improves robustness in physics-informed neural networks \\
\hline
\citep{jiang2023dynamic} & Dynamically reduces impact of noisy or hard-to-learn samples & Improves robustness to noise; enhances generalization & May underemphasize difficult samples & Achieves state-of-the-art performance on noisy datasets \\
\hline
\citep{song2023dynamic} & Uses an autoencoder to assign larger weights to difficult organs & Promotes consistent performance across organs; eliminates manual tuning & Adds computational overhead & Outperforms baselines and state-of-the-art methods on multiorgan segmentation \\
\hline
\citep{runda2020lstm} & Dynamically adjusts weights based on error rates observed after each training round & Effectively handles class imbalance; improves prediction accuracy & Requires additional computation for error rate calculations & Improves sentiment classification accuracy \\
\hline
\citep{roy2023margin} & Combines L-softmax with adaptive weighting based on inverse class frequencies & Addresses class imbalance; reduces overfitting & Requires careful tuning of margin parameters & Enhances performance on imbalanced datasets \\
\hline
\citep{groenendijk2021multi} & Dynamically adjusts weights based on coefficient of variation of each loss component & Eliminates manual tuning; adapts to training dynamics & Struggles with highly volatile loss components & Improves balanced performance across multiple tasks \\
\hline
\citep{maldonado2023owadapt} & Dynamically adjusts class weights using Ordered Weighted Averaging (OWA) operators & Versatile and easy to implement; addresses class imbalance & Requires tuning of aggregation parameters & Outperforms focal loss and cross-entropy on classification tasks \\
\hline
\citep{xiang2022self} & Dynamically adjusts weights of physics and boundary losses & Improves robustness in PINNs; reduces sensitivity to weight choices & Requires careful initialization of adaptive parameters & Enhances performance in physics-informed neural networks \\
\hline
\citep{heydari2019softadapt} & Dynamically updates weights of multi-part loss functions based on live performance metrics & Fast and easy to implement; compatible with any gradient-based optimizer & Sensitive to hyperparameter $\beta$ & Improves performance in multi-objective optimization tasks \\
\hline
Ours & Dynamically adjusts weights using Variance, MAD, and Bayesian-based methods & Eliminates manual tuning; Adapts to dynamic scenarios; Flexible strategies & Requires careful selection of weighting method & Achieves superior performance on breast ultrasound image segmentation

\end{tabular}
}
\end{table}

\section{Conclusion}

In this study, we introduced the DMF framework for adaptive multi-loss function optimization in deep learning, specifically applied to breast cancer segmentation. The framework allows for real-time adjustment of loss function weights, effectively addressing challenges such as class imbalance and varying task importance during training.

Our experiments on the BUSI, BUSC, and BUS datasets demonstrated improvements in segmentation performance, supported by various evaluation metrics, including Dice, IoU, and Precision. The incorporation of CB-Dice loss has shown potential in mitigating class imbalance issues. The DMF framework is not limited to medical imaging and can be applied to other deep learning tasks requiring multiple loss functions. By providing open-source code, we aim to enhance reproducibility and encourage further exploration of this approach.

While deep learning classification tasks are generally less complex and typically involve only a single loss function, segmentation presents greater challenges. In segmentation, multiple loss functions are often necessary to address pixel-wise predictions, boundary accuracy, and class imbalance. This makes it a more suitable domain for testing adaptive optimization frameworks like DMF. Therefore, our focus on segmentation tasks allows us to fully explore and demonstrate the effectiveness of the DMF framework in managing these complex tasks.

Although the proposed approach has proven to be effective, it may still have certain limitations that require further investigation across different tasks. The DMF framework's performance depends on the choice of weighting methods (Variance-based, MAD-based, or Bayesian-based), which yield varying results across datasets, indicating a lack of universal robustness. This sensitivity can be particularly challenging for large datasets, where selecting the appropriate weighting strategy for loss functions becomes critical. Therefore, it is necessary to choose a strategy based on the objective and the nature of the dataset, which underscores the need for deeper exploration to ensure optimal deployment in diverse scenarios.

For future work, we suggest investigating the DMF framework's scalability and its application in diverse machine learning scenarios. Previous observations may also be beneficial in limiting weight adjustments and exploring additional methods, such as exponential moving averages (EMA), for optimizing weight updates. Additionally, we suggest that future work could explore the use of an exhaustive grid search to further optimize hyperparameters, as this approach may yield improved model performance despite its computational demands.

\section*{Data Availability Statement}

\sloppy
The BUSI dataset can be accessed openly at \href{https://www.kaggle.com/datasets/sabahesaraki/breast-ultrasound-images-dataset}{https://www.kaggle.com/datasets/\\sabahesaraki/breast-ultrasound-images-dataset}. The BUSC dataset is available at \href{https://data.mendeley.com/datasets/vckdnhtw26/1}{https://data.mendeley.com/datasets/vckdnhtw26/1}, while the BUS Synthetic dataset can be found at \href{https://data.mendeley.com/datasets/r4phtn49r7/1}{https://data.mendeley.com/datasets/r4phtn49r7/1}.

\section*{Declaration of Competing Interest}

The authors declare that they have no known competing financial interests or personal relationships that could have appeared to influence the work reported in this paper.

\section*{Authorship Contribution Statement}

\textbf{Amin Golnari:} Conceptualization, Software, Investigation, Methodology, Visualization, Writing – Original Draft, Writing – Review \& Editing. \textbf{Mostafa Diba:} Software, Investigation, Writing – Original Draft, Writing – Review \& Editing.

\section*{Declaration of Generative AI in Scientific Writing}
During the preparation of this work, the authors used OpenAI's GPT-4o to enhance overall readability, improve language, and correct grammatical errors. After using this service, the authors reviewed and edited the content as needed and take full responsibility for the content of the publication.

\bibliographystyle{elsarticle-num} 
\bibliography{DynamicMemoryFusion}

\begin{thebibliography}{10}
\expandafter\ifx\csname url\endcsname\relax
  \def\url#1{\texttt{#1}}\fi
\expandafter\ifx\csname urlprefix\endcsname\relax\def\urlprefix{URL }\fi
\expandafter\ifx\csname href\endcsname\relax
  \def\href#1#2{#2} \def\path#1{#1}\fi

\bibitem{feng2024gbe}
J.~Feng, X.~Dong, S.~Chen, L.~Zhou, X.~Zheng, Gbe-net: Global boundary enhancement network for breast lesion segmentation in ultrasound images, Biomedical Signal Processing and Control 96 (2024) 106644.

\bibitem{wu2024mfmsnet}
R.~Wu, X.~Lu, Z.~Yao, Y.~Ma, Mfmsnet: A multi-frequency and multi-scale interactive cnn-transformer hybrid network for breast ultrasound image segmentation, Computers in Biology and Medicine 177 (2024) 108616.

\bibitem{zhu2024brain}
Z.~Zhu, Z.~Wang, G.~Qi, N.~Mazur, P.~Yang, Y.~Liu, Brain tumor segmentation in mri with multi-modality spatial information enhancement and boundary shape correction, Pattern Recognition 153 (2024) 110553.

\bibitem{zhu2024lightweight}
Z.~Zhu, K.~Yu, G.~Qi, B.~Cong, Y.~Li, Z.~Li, X.~Gao, Lightweight medical image segmentation network with multi-scale feature-guided fusion, Computers in Biology and Medicine 182 (2024) 109204.

\bibitem{zhu2024sparse}
Z.~Zhu, M.~Sun, G.~Qi, Y.~Li, X.~Gao, Y.~Liu, Sparse dynamic volume transunet with multi-level edge fusion for brain tumor segmentation, Computers in Biology and Medicine (2024) 108284.

\bibitem{zhu2023brain}
Z.~Zhu, X.~He, G.~Qi, Y.~Li, B.~Cong, Y.~Liu, Brain tumor segmentation based on the fusion of deep semantics and edge information in multimodal mri, Information Fusion 91 (2023) 376--387.

\bibitem{zhu2025dual}
Z.~Zhu, Z.~Zhang, G.~Qi, Y.~Li, Y.~Li, L.~Mu, A dual-branch network for ultrasound image segmentation, Biomedical Signal Processing and Control 103 (2025) 107368.

\bibitem{heydari2019softadapt}
A.~A. Heydari, C.~A. Thompson, A.~Mehmood, Softadapt: Techniques for adaptive loss weighting of neural networks with multi-part loss functions, arXiv preprint arXiv:1912.12355 (2019).

\bibitem{fernando2021dynamically}
K.~R.~M. Fernando, C.~P. Tsokos, Dynamically weighted balanced loss: class imbalanced learning and confidence calibration of deep neural networks, IEEE Transactions on Neural Networks and Learning Systems 33~(7) (2021) 2940--2951.

\bibitem{guermazi2022dynamically}
B.~Guermazi, R.~Ksantini, N.~Khan, A dynamically weighted loss function for unsupervised image segmentation, in: 2022 IEEE Eighth International Conference on Multimedia Big Data (BigMM), IEEE, 2022, pp. 73--78.

\bibitem{song2023dynamic}
Y.~Song, J.~Y.-C. Teoh, K.-S. Choi, J.~Qin, Dynamic loss weighting for multiorgan segmentation in medical images, IEEE transactions on neural networks and learning systems (2023).

\bibitem{erdogan2024fuseformer}
A.~Erdogan, E.~Akagunduz, Fuseformer: A transformer for visual and thermal image fusion, arXiv preprint arXiv:2402.00971 (2024).

\bibitem{zhao2024s2f}
Y.~Zhao, Y.~Xia, Y.~Ding, Y.~Liu, S.~Liu, H.~Wang, S2f-net: Shared-specific fusion network for infrared and visible image fusion, in: Proceedings of the 2024 International Conference on Multimedia Retrieval, 2024, pp. 497--505.

\bibitem{ou2021fusion}
W.-F. Ou, L.-M. Po, C.~Zhou, Y.~A.~U. Rehman, P.-F. Xian, Y.-J. Zhang, Fusion loss and inter-class data augmentation for deep finger vein feature learning, Expert Systems with Applications 171 (2021) 114584.

\bibitem{tasnim2023cam}
J.~Tasnim, M.~K. Hasan, Cam-qus guided self-tuning modular cnns with multi-loss functions for fully automated breast lesion classification in ultrasound images, Physics in Medicine \& Biology 69~(1) (2023) 015018.

\bibitem{jiang2023dynamic}
S.~Jiang, J.~Li, J.~Zhang, Y.~Wang, T.~Xu, Dynamic loss for robust learning, IEEE Transactions on Pattern Analysis and Machine Intelligence (2023).

\bibitem{jesson2017brain}
A.~Jesson, T.~Arbel, Brain tumor segmentation using a 3d fcn with multi-scale loss, in: International MICCAI Brainlesion Workshop, Springer, 2017, pp. 392--402.

\bibitem{sharifzadeh2023phase}
M.~Sharifzadeh, H.~Benali, H.~Rivaz, Phase aberration correction without reference data: An adaptive mixed loss deep learning approach, arXiv preprint arXiv:2303.05747 (2023).

\bibitem{lv2024ssdfusion}
Q.~Lv, R.~Yang, Y.~Chen, Z.~Zhou, C.~Zhang, S.~Liu, Ssdfusion: A semantic segmentation driven framework for infrared and visible image fusion (2024).

\bibitem{gao2019multi}
Q.~Gao, Q.~Peng, S.~Xuan, K.~Xiong, Multi-loss function fusion for face recognition based on the convolutional neural network, in: 2019 12th International Congress on Image and Signal Processing, BioMedical Engineering and Informatics (CISP-BMEI), IEEE, 2019, pp. 1--6.

\bibitem{zhao2024learning}
Y.~Zhao, X.~Shen, J.~Chen, W.~Qian, L.~Sang, H.~Ma, Learning active contour models based on self-attention for breast ultrasound image segmentation, Biomedical Signal Processing and Control 89 (2024) 105816.

\bibitem{zhao2024loss}
Y.~Zhao, X.~Shen, J.~Chen, W.~Qian, H.~Ma, L.~Sang, loss for low-contrast medical image segmentation, Machine Learning: Science and Technology 5~(1) (2024) 015013.

\bibitem{chao2020multi}
F.-Y. Chao, L.~Zhang, W.~Hamidouche, O.~D{\'e}forges, A multi-fov viewport-based visual saliency model using adaptive weighting losses for 360$^\circ$ images, IEEE Transactions on Multimedia 23 (2020) 1811--1826.

\bibitem{sushma2024aapfc}
B.~Sushma, A.~Pulikala, Aapfc-busnet: Hierarchical encoder--decoder based cnn with attention aggregation pyramid feature clustering for breast ultrasound image lesion segmentation, Biomedical Signal Processing and Control 91 (2024) 105969.

\bibitem{mang2024dynamic}
C.~Mang, A.~Tahmasebimoradi, D.~Danan, M.~Yagoubi, A dynamic weighted loss function for enhancing the performance of neural networks, in: 16th World Congress on Computational Mechanics (WCCM), 2024.

\bibitem{lu2019dynamic}
S.~Lu, F.~Gao, C.~Piao, Y.~Ma, Dynamic weighted cross entropy for semantic segmentation with extremely imbalanced data, in: 2019 International conference on artificial intelligence and advanced manufacturing (AIAM), IEEE, 2019, pp. 230--233.

\bibitem{maldonado2023owadapt}
S.~Maldonado, C.~Vairetti, K.~Jara, M.~Carrasco, J.~L{\'o}pez, Owadapt: An adaptive loss function for deep learning using owa operators, Knowledge-Based Systems 280 (2023) 111022.

\bibitem{runda2020lstm}
Y.~Runda, X.~Yan, Z.~Mingfang, Z.~Li, W.~Hongbin, Lstm sentiment classification model based on one kind of dynamic loss weighting function, in: 2020 7th International Conference on Information Science and Control Engineering (ICISCE), IEEE, 2020, pp. 819--824.

\bibitem{roy2023margin}
D.~Roy, R.~Pramanik, R.~Sarkar, Margin-aware adaptive-weighted-loss for deep learning based imbalanced data classification, IEEE Transactions on Artificial Intelligence 5~(2) (2023) 776--785.

\bibitem{groenendijk2021multi}
R.~Groenendijk, S.~Karaoglu, T.~Gevers, T.~Mensink, Multi-loss weighting with coefficient of variations, in: Proceedings of the IEEE/CVF winter conference on applications of computer vision, 2021, pp. 1469--1478.

\bibitem{ocampo2024adaptive}
D.~Ocampo, D.~Posso, R.~Namakian, W.~Gao, Adaptive loss weighting for machine learning interatomic potentials, Computational Materials Science 244 (2024) 113155.

\bibitem{xiang2022self}
Z.~Xiang, W.~Peng, X.~Liu, W.~Yao, Self-adaptive loss balanced physics-informed neural networks, Neurocomputing 496 (2022) 11--34.

\bibitem{barron2019general}
J.~T. Barron, A general and adaptive robust loss function, in: Proceedings of the IEEE/CVF conference on computer vision and pattern recognition, 2019, pp. 4331--4339.

\bibitem{mrad2021machine}
I.~Mrad, R.~Hamila, A.~Erbad, T.~Hamid, R.~Mazhar, N.~Al-Emadi, Machine learning screening of covid-19 patients based on x-ray images for imbalanced classes, in: 2021 9th European Workshop on Visual Information Processing (EUVIP), IEEE, 2021, pp. 1--6.

\bibitem{al2020dataset}
W.~Al-Dhabyani, M.~Gomaa, H.~Khaled, A.~Fahmy, Dataset of breast ultrasound images, Data in brief 28 (2020) 104863.

\bibitem{iqbal2023unet}
A.~Iqbal, M.~Sharif, Unet: A semi-supervised method for segmentation of breast tumor images using a u-shaped pyramid-dilated network, Expert Systems with Applications 221 (2023) 119718.

\bibitem{rodrigues2017breast}
P.~S. Rodrigues, Breast ultrasound image, Mendeley Data 1~(10) (2017).

\bibitem{balocco2010srbf}
S.~Balocco, C.~Gatta, O.~Pujol, J.~Mauri, P.~Radeva, Srbf: Speckle reducing bilateral filtering, Ultrasound in medicine \& biology 36~(8) (2010) 1353--1363.

\bibitem{golnari2024probabilistic}
A.~Golnari, M.~H. Komeili, Z.~Azizi, Probabilistic deep learning and transfer learning for robust cryptocurrency price prediction, Expert Systems with Applications 255 (2024) 124404.

\end{thebibliography}

\end{document}